\documentclass{article}

\usepackage[final, nonatbib]{neurips_2020}

\usepackage[utf8]{inputenc} 
\usepackage[T1]{fontenc}   
\usepackage{hyperref}  
\usepackage{url}          
\usepackage{booktabs}     
\usepackage{amsfonts}       
\usepackage{nicefrac}       
\usepackage{microtype}      
\usepackage{todonotes}
\usepackage{comment}
\usepackage{amsthm}
\usepackage{amsmath, amssymb}

\usepackage[linesnumbered,ruled]{algorithm2e}
\usepackage{algorithmic, float}
\usepackage{subfigure}

\newcommand{\less}{\leqslant}
\newcommand{\gre}{\geqslant}

\newcommand{\argmin}{\mathop{\rm argmin}}

\newcommand{\defn}{\ensuremath{: \, =}}
\newcommand{\real}{\mathbb{R}}
\newcommand{\Exs}{\ensuremath{\mathbb{E}}}
\newcommand{\xstar}{x^*}

\newcommand{\wtilde}{\widetilde}

\newtheorem{remark}{Remark}[section]
\newtheorem{theo}{Theorem}[section]

\newtheorem{lem}{Lemma}[section]
\newtheorem{prop}{Proposition}[section]
\newtheorem{cor}{Corollary}[section]
\theoremstyle{definition} 
\newtheorem{nota}{Notation}[section]
\newtheorem{de}{Definition}[section]
\newtheorem{exa}{Example}[section]
\newtheorem{as}{Assumption}[section]
\newtheorem{alg}{Algorithm}[section]
\newcommand{\btheo}{\begin{theo}}
\newcommand{\bde}{\begin{de}}
\newcommand{\ble}{\begin{lem}}
\newcommand{\bpr}{\begin{prop}}
\newcommand{\bno}{\begin{nota}}
\newcommand{\bex}{\begin{exa}}
\newcommand{\bcor}{\begin{cor}}
\newcommand{\spro}{\begin{proof}}
\newcommand{\bas}{\begin{as}}
\newcommand{\balg}{\begin{alg}}
\newcommand{\etheo}{\end{theo}}
\newcommand{\ede}{\end{de}}
\newcommand{\ele}{\end{lem}}
\newcommand{\epr}{\end{prop}}
\newcommand{\eno}{\end{nota}}
\newcommand{\eex}{\end{exa}}
\newcommand{\ecor}{\end{cor}}
\newcommand{\fpro}{\end{proof}}
\newcommand{\eas}{\end{as}}
\newcommand{\ealg}{\end{alg}}
\newtheorem{theos}{Theorem}
\newtheorem{props}{Proposition}
\newtheorem{lems}{Lemma}
\newtheorem{cors}{Corollary}
\theoremstyle{definition}
\newtheorem{exas}{Example}
\newtheorem{algs}{Algorithm}
\newtheorem{asss}{Assumption}
\newtheorem{defns}{Definition}
\newcommand{\btheos}{\begin{theos}}
\newcommand{\etheos}{\end{theos}}
\newcommand{\bprops}{\begin{props}}
\newcommand{\eprops}{\end{props}}
\newcommand{\bdes}{\begin{defns}}
\newcommand{\edes}{\end{defns}}
\newcommand{\blems}{\begin{lems}}
\newcommand{\elems}{\end{lems}}
\newcommand{\bcors}{\begin{cors}}
\newcommand{\ecors}{\end{cors}}
\newcommand{\bexs}{\begin{exas}}
\newcommand{\eexs}{\end{exas}}
\newcommand{\balgs}{\begin{algs}}
\newcommand{\ealgs}{\end{algs}}
\newcommand{\bass}{\begin{asss}}
\newcommand{\eass}{\end{asss}}

\title{Effective Dimension Adaptive Sketching Methods for Faster Regularized Least-Squares Optimization}

\author{
  Jonathan Lacotte\\
  Department of Electrical Engineering\\
  Stanford University\\
  \texttt{lacotte@stanford.edu} \\
  \And
  Mert Pilanci \\
  Department of Electrical Engineering \\
  Stanford University \\
  \texttt{pilanci@stanford.edu} \\
}

\begin{document}

\maketitle

\begin{abstract}
We propose a new randomized algorithm for solving L2-regularized least-squares problems based on sketching. We consider two of the most popular random embeddings, namely, Gaussian embeddings and the Subsampled Randomized Hadamard Transform (SRHT). While current randomized solvers for least-squares optimization prescribe an embedding dimension at least greater than the data dimension, we show that the embedding dimension can be reduced to the effective dimension of the optimization problem, and still preserve high-probability convergence guarantees. In this regard, we derive sharp matrix deviation inequalities over ellipsoids for both Gaussian and SRHT embeddings. Specifically, we improve on the constant of a classical Gaussian concentration bound whereas, for SRHT embeddings, our deviation inequality involves a novel technical approach. Leveraging these bounds, we are able to design a practical and adaptive algorithm which does not require to know the effective dimension beforehand. Our method starts with an initial embedding dimension equal to 1 and, over iterations, increases the embedding dimension up to the effective one at most. Hence, our algorithm improves the state-of-the-art computational complexity for solving regularized least-squares problems. Further, we show numerically that it outperforms standard iterative solvers such as the conjugate gradient method and its pre-conditioned version on several standard machine learning datasets.
\end{abstract}

% TODO

% 3) in the complexity analysis part, give the complexity for A a sparse matrix. mention whether we could extend our analysis to sparse sketching matrices.

% 4) in the description of the algorithm, give specific values for eta and rho.

\section{Introduction}
\label{SectionIntroduction}

We study the performance of a randomized method, namely, the Hessian sketch~\cite{pilanci2016iterative}, in the context of regularized least-squares problems,
\begin{align}
\label{EqnMain}
    x^* \defn \argmin_{x \in \real^d} \left\{f(x) \defn \frac{1}{2} {\|Ax-b\|}_2^2 + \frac{\nu^2}{2} {\|x\|}_2^2\right\}\,,
\end{align}
where $A \in \real^{n \times d}$ is a data matrix and $b \in \real^n$ is a vector of observations. For clarity purposes and without loss of generality (by considering instead the dual problem of~\eqref{EqnMain}), we make the assumption that the problem is over-determined, i.e., $n \gre d$ and that $\text{rank}(A) = d$.

The regularized solution $x^*$ can be obtained using direct methods which have computational complexity $\mathcal{O}(nd^2)$. In the large-scale setting $n, d \gg 1$, this is prohibitively large. A linear dependence $\wtilde{\mathcal{O}}(nd)$ is preferable and this can be obtained by using first-order iterative solvers~\cite{grewal2014kalman} such as the conjugate gradient method (CG) for which the per-iteration complexity scales as $\mathcal{O}(nd)$. Using the standard prediction (semi)-norm error $\frac{1}{2}\|\overline A(\widetilde x - x^*)\|_2^2$ where $\overline A \defn \begin{bmatrix} A \\ \nu I_d \end{bmatrix}$ as the evaluation criterion for an estimator $\widetilde x$, these iterative methods have time complexity which usually scales proportionally to the condition number $\kappa$ of $\overline A$ (or $\sqrt{\kappa}$ with acceleration) in order to find a solution $\widetilde x$ with acceptable accuracy. This also becomes prohibitively large when $\kappa \gg 1$. Besides the computational complexity, the number of iterations of an iterative solver is also a relevant performance metric in the large-scale setting, as distributed computation may be necessary at each iteration. In this regard, randomized preconditioning methods~\cite{rokhlin2008fast, avron2010blendenpik, meng2014lsrn} involve using a random matrix $S \in \real^{m \times n}$ with $m \ll n$ to project the data $A$, and then improve the condition number of $A$ based on a spectral decomposition of $SA$. On the other hand, the iterative Hessian sketch (IHS) introduced by~\cite{pilanci2016iterative} and considered in~\cite{ozaslan2019iterative, lacotte2019faster, lacotte2020optimal, ozaslan2019regularized, pilanci2017newton} addresses the conditioning issue differently. Given $x_0, x_1 \in \real^d$, it uses a pre-conditioned Heavy-ball update with step size $\mu$ and momentum parameter $\beta$, given by
\begin{align}
\label{EqnPolyak}
    x_{t+1} = x_{t} - \mu H_{S}^{-1} \nabla f(x_t) + \beta (x_t\!-\!x_{t-1})
\end{align}
where the Hessian $H \defn \overline A^\top \overline A$ of $f(x)$ is approximated by $H_{\overline{S}} = \overline{A}^\top \overline{S}^\top \overline{S}\, \overline A$ and $\overline{S}$ is a sketching matrix. We refer to the update~\eqref{EqnPolyak} as the \emph{Polyak-IHS method}, and, in the absence of acceleration ($\beta = 0$), we call it the \emph{gradient-IHS method}. In contrast to preconditioning methods~\cite{rokhlin2008fast, avron2010blendenpik, meng2014lsrn}, the IHS does not need to pay the full cost $\mathcal{O}(md\min\{m,d\})$ for decomposing the matrix $SA$. Although solving exactly the linear system $H_{\overline{S}} \cdot z = \nabla f(x_t)$ also takes time $\mathcal{O}(md\min\{m,d\})$, approximate solving (using for instance CG) is also efficient and faster in practice~\cite{ozaslan2019regularized, ozaslan2019iterative}.

The choice of the sketching matrix $S$ is critical for statistical and computational performances. A classical sketch is a matrix $S$ with independent and identically distributed (i.i.d.) Gaussian entries $\mathcal{N}(0,m^{-1})$ for which forming $SA$ requires in general $\mathcal{O}(mnd)$ basic operations (using classical matrix multiplication). On the other hand, it has been observed~\cite{mahoney2011randomized, drineas2016randnla} and also formally proved~\cite{dobriban2019asymptotics, lacotte2020optimal} in several contexts that random projections with i.i.d.~entries degrade the performance of the approximate solution compared to orthogonal projections. In this regard, the SRHT~\cite{ailon2006approximate} is an orthogonal embedding for which the sketch $SA$ can be formed in $\mathcal{O}(nd \log m)$ time, and this is much faster than Gaussian projections. Consequently, along with the statistical benefits of orthogonal projections, this suggests to use the SRHT as a reference point for comparing sketching algorithms.

In the context of \emph{unregularized} least-squares problems ($\nu=0$),~\cite{lacotte2019faster} showed that the error $\frac{1}{2}\|A(x_t - x^*)\|_2^2$ of the Polyak-IHS method is smaller than $(d/m)^t$ for both Gaussian and SRHT matrices provided that $m \approx d$. More recently, it has been shown in~\cite{lacotte2020optimal} that the scaling $(d/m)^t$ is exact for Gaussian embeddings in the asymptotic regime where we let the relevant dimensions go to infinity, whereas the exact scaling for the SRHT is slightly smaller than $(d/m)^t$. 

In the \emph{regularized} case ($\nu > 0$), more relevant than the matrix rank is the \emph{effective dimension} $d_e \defn \text{trace}(A (A^\top A + \nu^2 I_d)^{-1}A^\top)$ which always satisfies $d_e \less d$, and it is significantly smaller than $d$ when the matrix $A$ has a fast spectral decay. It has been shown in~\cite{ozaslan2019regularized} that one can pick $m \approx d_e$ and achieve the error rate $(d_e/m)^t$ by using the well-structured approximate Hessian
\begin{align}
\label{EqnApproximateHessian}
    H_{S} \defn A^\top S^\top S A + \nu^2 \cdot I\,.
\end{align}
Further, with $m \approx d_e$ instead of $m \approx d$, the linear system $H_S\cdot z = \nabla f(x_t)$ can be solved in time $\mathcal{O}(d_e^2 d)$ instead of $\mathcal{O}(d^3)$ by computing and caching a factorization of $SA$ and then using the Woodbury matrix identity~\cite{hager1989updating} to invert $H_S$.

However, it is necessary to estimate $d_e$ (which is usually unknown) to be able to pick $m \approx d_e$ and then achieve these computational and memory space savings. The randomized technique proposed by~\cite{avron2017sharper} can be used to estimate $d_e$, but under the restrictive assumption that $d_e$ is very small (e.g., see Theorem 60 in~\cite{avron2017sharper}). In~\cite{ozaslan2019regularized}, the authors propose to use a heuristic Hutchinson-type trace estimator~\cite{avron2011randomized} and do not provide any guarantee on the estimation accuracy of $d_e$. Consequently, our \emph{main goal} in this paper is to design an adaptive algorithm which does not require the knowledge of $d_e$, but is still able to use a sketch size $m \lesssim d_e$ and achieve an error rate $(d_e/m)^t$.

State-of-the-art randomized preconditioning methods~\cite{rokhlin2008fast, avron2010blendenpik, meng2014lsrn} prescribe to use $m$ proportional to $d$ in the context of unregularized least-squares problems. Since it appears non-trivial to adapt and analyze these methods to the regularized case with sketch sizes $m \approx d_e$, nor to design an adaptive scheme which does not require the knowledge of $d_e$, we focus our attention to the Polyak-IHS method in this work.

\subsection{Notations}

We denote by $\|z\|$ or $\|z\|_2$ the Euclidean norm of a vector $z$, $\|M\|_2$ the operator norm of a matrix $M$ and $\|M\|_F$ its Frobenius norm. 

We introduce the diagonal matrix $D \defn \textrm{diag}\!\left(\frac{\sigma_1}{\sqrt{\sigma_1^2+\nu^2}},\dots,\frac{\sigma_d}{\sqrt{\sigma_d^2+\nu^2}}\right)$ where $\sigma_1 \gre \dots \gre \sigma_d$ are the singular values of the matrix $A$. We define the effective dimension as $d_e \defn \frac{\|D\|_F^2}{\|D\|_2^2}$. We denote by $U \in \real^{n \times d}$ a matrix of left singular vectors of $A$ and by $\overline U \in \real^{(n+d) \times d}$ a matrix of left singular vectors of $\overline A\defn \begin{bmatrix} A \\ \nu \cdot I_d \end{bmatrix}$. 

Given a sequence of iterates $\{x_t\}$, we define its error at time $t$ as $\delta_t \defn \frac{1}{2}\|\overline A(x_t-\xstar)\|^2$.

For a sketching matrix $S \in \real^{m \times n}$, we denote the approximate Hessian $H_S \defn A^\top S^\top S A + \nu^2 I_d$, and the exact Hessian $H \defn \overline A^\top \overline A$. Critical to our convergence analysis is the matrix $C_S \defn D(U^\top S^\top S U - I_d) D + I_d$.

\subsection{Overview of our contributions}

Our main contribution is to propose an iterative method that does not require the knowledge of $d_e$, and is still able to achieve the error rate $\mathcal{O}\left((d_e/m)^t\right)$. Our method is initialized with an arbitrary $m$ (e.g, $m=1$) and, at each iteration of the Polyak-IHS update~\eqref{EqnPolyak}, it uses an \emph{improvement criterion} to decide whether it should increase $m$ or not. We prove that the \emph{adaptive} sketch size satisfies at each iteration $m \lesssim d_e$ and that our algorithm improves on the state-of-the-art computational complexity for solving regularized least-squares problems.

Our algorithmic parameters and improvement criterion depend on the extreme eigenvalues of $C_S$, and it is then critical for optimal performance to have a sharp estimation of these. For Gaussian embeddings, we provide a sharper constant for well-known Gaussian concentration bounds~\cite{koltchinskii2017concentration}. Our constant is tight in a worst-case sense, and our analysis is based on a recent extension~\cite{thrampoulidis2014tight} of Gordon's min-max theorem~\cite{gordon1985some}. In the SRHT case, although similar concentration bounds were already obtained (e.g., see Theorem~1 in~\cite{cohen2016optimal}), we provide a novel technical approach which generalizes the classical results and analysis proposed in~\cite{tropp2011improved}.

We evaluate numerically our adaptive algorithm on several standard datasets. We consider two settings: (i) the regularization parameter $\nu$ is fixed; (ii) one aims to compute the several solutions along a regularization path. The latter setting is more relevant to many practical applications~\cite{vogel2002computational, hastie2009elements} where estimating a proper regularization parameter is essential. In both cases, we show empirically that our method is faster than the standard conjugate gradient method and one of the state-of-the-art randomized preconditioning methods~\cite{rokhlin2008fast}. 

Finally, we address the underdetermined case $d \gre n$. By considering the dual of~\eqref{EqnMain} which is itself an overdetermined regularized least-squares problem, we show that our adaptive algorithm and theoretical guarantees apply to this setting. We defer the presentation of these results to Appendix~\ref{SectionUnderdetermined}.

\subsection{Other related work}

Another class of sketch-and-solve algorithms project both $A$ and $b$, and then computes $\widetilde x \defn \argmin_x \frac{1}{2} \|SA x - Sb\|_2^2 + \frac{\lambda}{2} \|x\|_2^2$ (see e.g. \cite{drineas2016randnla,pilanci2015randomized, pilanci2016fast, sridhar2020lower, derezinski2020debiasing, bartan2020distributedaveraging,bartan2020distributedsketching}). In~\cite{avron2017sharper}, the authors showed that for $m \approx d_e / \varepsilon$, the estimate $\widetilde x$ satisfies $f(\widetilde x) \less (1+\varepsilon) f(x^*)$. This can result in large $m$ for even medium accuracy, whereas our method yields an $\varepsilon$-approximate solution with $m \approx d_e$ under the mild requirement that the number of iterations $T$ satisfies $T \approx \log(1/\varepsilon)$. Further, the effective dimension can be efficiently estimated only in limited settings (e.g., see Theorem 60 in~\cite{avron2017sharper}). Closely related to our work is the iterative method proposed by~\cite{chowdhury2018iterative} for solving underdetermined ridge regression problems. It involves a similar approximation of the Hessian $\overline A^\top \overline A$ by $H_S$, where the sketch size $m$ depends on the effective dimension $d_e$ as opposed to the data dimension $d$. However their proposed method also requires prior knowledge or estimation of $d_e$. Several sketch-and-solve algorithms~\cite{chen2015fast, wang2017sketched} for ridge regression were not analyzed in terms of $d_e$ but $d$. In the context of kernel ridge regression, it was shown that Nystrom approximations of kernel matrices have performance guarantees for sketch sizes proportional to the effective dimension~\cite{bach2013sharp, alaoui2015fast, cohen2017input}.

Other versions of the IHS have been proposed in the literature, especially in the context of unregularized least-squares. A fundamentally different version uses the same update~\eqref{EqnPolyak} but with refreshed sketching matrices, i.e., a new matrix $S$ is sampled at each iteration and independently of the previous ones, and the approximate Hessian $H_S$ is re-computed. Surprisingly, refreshing embeddings does not improve on using a fixed embedding: it results in the same convergence rate in the Gaussian case~\cite{lacotte2019faster, lacotte2020optimal} and in a slower convergence rate in the SRHT case~\cite{lacotte2020optimal}.

\section{Preliminaries}
\label{SectionPreliminaries}

We provide deterministic convergence guarantees for the Polyak- and gradient-IHS methods, and we relate the convergence rates to the extreme eigenvalues of the matrix $C_S$.

Let $S \in \real^{m \times d}$ be any sketching matrix with arbitrary sketch size $m$, and denote by $\gamma_1$ (resp.~$\gamma_d$) the largest (resp.~smallest) eigenvalue of $C_S$. Since the matrix $D^\top U^\top S^\top S U D$ is positive semi-definite and $\|D\|_2 < 1$, it holds that $C_S$ is positive definite. Given two real numbers $\Lambda > \lambda > 0$, we define the $S$-measurable event $\mathcal{E}_S \defn \left\{ \lambda \less \gamma_d \less \gamma_1 \less \Lambda \right\}$. The proofs of the two next results are based on standard analyses of gradient methods~\cite{polyak1964some}, and they are deferred to Appendix~\ref{ProofTheoremDeterministicRates}.
\btheos
\label{TheoremDeterministicRateGradientMethod}
Consider the step size $\mu_\text{gd}(\lambda,\Lambda) \defn 2 / (\frac{1}{\lambda} + \frac{1}{\Lambda})$. Then, conditional on $\mathcal{E}_S$, the gradient-IHS method satisfies at each iteration
\begin{align}
\label{EqnDetermisticErrorGradientMethod}
    \frac{\delta_{t+1}}{\delta_t} \less c_\text{gd}(\lambda, \Lambda)\,,\,\quad \text{where} \quad c_\text{gd}(\lambda, \Lambda) \defn \left(\frac{\Lambda-\lambda}{\Lambda+\lambda}\right)^2\,.
\end{align}
\etheos

\btheos
\label{TheoremDeterministicRatePolyak}
Consider the step size $\mu_p(\lambda, \Lambda) \defn 4 / (\frac{1}{\sqrt{\lambda}} + \frac{1}{\sqrt{\Lambda}})^{2}$ and momentum parameter $\beta_p(\lambda, \Lambda) \defn \left(\frac{\sqrt{\Lambda}-\sqrt{\lambda}}{\sqrt{\Lambda} + \sqrt{\lambda}}\right)^2$. Then, conditional on $\mathcal{E}_S$, the Polyak-IHS satisfies 
\begin{align}
\label{EqnDetermisticErrorPolyak}
    \limsup_{t \to \infty} \,\, \left(\frac{\delta_t}{\delta_0}\right)^{\frac{1}{t}} \less c_p(\lambda, \Lambda)\,,\,\quad \text{where} \quad c_p(\lambda, \Lambda) \defn \left(\frac{\sqrt{\Lambda}-\sqrt{\lambda}}{\sqrt{\Lambda} + \sqrt{\lambda}}\right)^2\,.
\end{align}
\etheos
The above rates $c_\text{gd}(\lambda, \Lambda)$ and $c_p(\lambda, \Lambda)$ will play a critical role in the design of our adaptive method. For the gradient-IHS method, it should be noted that we are able to monitor the improvement ratio between two consecutive iterates. However, for the Polyak-IHS method, we only obtain an asymptotic guarantee as $t \to +\infty$. This standard result regarding the Heavy-ball method~\cite{polyak1964some} essentially follows from the fact that the iterates obey a non-symmetric linear dynamical system so that, according to Gelfand's formula, the spectral and operator norms of this linear system only coincide asymptotically.

\section{Sharp convergence rates for Gaussian and SRHT embeddings}

According to Theorems~\ref{TheoremDeterministicRateGradientMethod} and~\ref{TheoremDeterministicRatePolyak}, we need sharp estimates of the extreme eigenvalues of $C_S$ in order to pick optimal parameters for the Polyak- and gradient-IHS methods.

\subsection{The Gaussian case}
\label{SectionGaussianBounds}
We provide a concentration bound on the edge eigenvalues $\gamma_1$ and $\gamma_d$ of the matrix $C_S$ in terms of the aspect ratio $\frac{d_e}{m}$. Our analysis is based on a generalized Gordon's Gaussian comparison theorem~\cite{gordon1985some,thrampoulidis2014tight} and it provides sharper constants than existing results. We defer the proof to Appendix~\ref{ProofGaussianConcentrationEllipsoids}. 
\btheos
\label{TheoremGaussianConcentrationEllipsoids}
Let $\rho, \eta > 0$ be some parameters, and $S \in \real^{m \times n}$ be a Gaussian embedding with $m \gre \frac{d_e}{\rho}$. Then, it holds with probability at least $1- 8 e^{-m \rho \eta/2}$ that 
\begin{align}
\label{eqngaussiancontrationbounds}
    \begin{cases}
    & \gamma_1 \less 1 - \|D\|^2 + \|D\|_2^2  (1+\sqrt{c_\eta \rho})^2\\
    & \gamma_d \gre 1-\|D\|_2^2 + \|D\|_2^2 (1-\sqrt{c_\eta \rho})^2\,,\qquad \text{provided that} \quad \rho \in (0,0.18], \, \eta \in (0.01]\,. 
    \end{cases}
\end{align}
where $c_\eta \defn (1+3\sqrt{\eta})^2$.
\etheos 
The lower\footnote{For the lower bound, we use the restrictions $\rho \less 0.18$ and $\eta \less 0.01$ for the sake of having simple expressions, while covering a range of values useful in practice. However, similar lower bounds hold for any $\rho \in (0,1)$ and small enough $\eta$.} and upper bounds~\eqref{eqngaussiancontrationbounds} are respectively increasing and decreasing in $\|D\|_2$, so that one can replace the potentially unknown quantity $\|D\|_2$ by $1$ as follows.
\bde[Practical parameters for Gaussian embeddings]
\label{DefTargetsGaussian}
Given $\rho \less 0.18$ and $\eta \less 0.01$, we define the bounds $\lambda_{\rho,\eta} \defn (1-\sqrt{c_\eta \rho})^2$ and $\Lambda_{\rho,\eta} \defn (1+\sqrt{c_\eta \rho})^2$ where $c_\eta = (1+3\sqrt{\eta})^2$. We denote the corresponding algorithmic parameters by $\mu_\text{gd}(\rho,\eta) \defn \mu_\text{gd}(\lambda_{\rho,\eta}, \Lambda_{\rho,\eta})$, $\mu_\text{p}(\rho,\eta) \defn \mu_\text{p}(\lambda_{\rho,\eta}, \Lambda_{\rho,\eta})$ and $\beta_\text{p}(\rho,\eta) \defn \beta_\text{p}(\lambda_{\rho,\eta}, \Lambda_{\rho,\eta})$, and the corresponding convergence rates $c_\text{gd}(\rho,\eta) \defn c_{\text{gd}}( \lambda_{\rho,\eta},\Lambda_{\rho,\eta})$ and $c_\text{p}(\rho,\eta) \defn c_{\text{p}}(\lambda_{\rho,\eta},\Lambda_{\rho,\eta})$.
\ede
According to Theorems~\ref{TheoremDeterministicRateGradientMethod} and~\ref{TheoremDeterministicRatePolyak}, the closer the bounds on $\gamma_1$ and $\gamma_d$ to $1$, the faster the convergence rates of the Gradient- and Polyak-IHS updates. Consequently, one needs to pick both $\rho$ and $\eta$ small. However, this trades off, on the one hand, with a larger sketch size $m$ (i.e., higher computational costs) and, on the other hand, with a weaker probabilistic guarantee. For instance, suppose that $\rho \approx 0.1$ and $\eta \approx 0.01$ are fixed. This results in $m \gtrsim 10^3$ to get low failure probability $e^{-m \rho \eta}$. Such a choice of the sketch size is particularly relevant when $d_e / \rho \gtrsim 10^3$, i.e., $d_e \gtrsim 10^2$, and $\min\{n,d\} \gg 10^3$. On the other hand, in the very small $d_e$ regime, it is harder to keep $m$ close to the \emph{target sketch size} $d_e/\rho$. Since our sketching-based method relies on measure concentration phenomena, this should be expected. 
\begin{remark}
Letting $d_e, m \to +\infty$ while keeping the aspect ratio $\rho \defn \frac{d_e}{m}$ fixed and taking $\eta \sim 1/\sqrt{m}$, our bounds~\eqref{eqngaussiancontrationbounds} converge to the respective limits $1-\|D\|_2^2 + \|D\|_2^2 (1-\sqrt{\rho})^2$ and $1-\|D\|_2^2 + \|D\|_2^2 (1+\sqrt{\rho})^2$. When $D = \|D\|_2 \cdot I_d$, these limits are exact as they correspond to the edges of the support of the Marchenko-Pastur distribution~\cite{marchenko1967distribution}, so that our bounds are tight in a worst-case sense. Further, we have that $\|C_S - I_d\|_2 \less \|D\|_2^2 \left(2\sqrt{\rho} + \rho\right) (1+4\,m^{-\frac{1}{4}})$ with probability at least $1-8 e^{-\frac{\sqrt{m}\rho}{32}}$, whereas standard Gaussian concentration bounds (e.g., see~\cite{koltchinskii2017concentration}) states that $\|C_S - I_d\|_2 \less \|D\|_2^2 \left(2\sqrt{\rho}+\rho\right) (1+c_0)$ with high probability for some universal constant $c_0 > 0$. In contrast, our factor $(1+4\,m^{-\frac{1}{4}})$ is asymptotically sharper. 
\end{remark}

\subsection{The SRHT case}
\label{SectionSRHTBounds}

We provide a concentration bound in terms of the aspect ratio $C(n,d_e) \cdot \frac{d_e \log(d_e)}{m}$ where we introduced the oversampling factor $C(n,d_e) \defn \frac{16}{3} (1 + \sqrt{\frac{8 \log(d_e n)}{d_e}})^2$. Under the mild requirement $d_e \gtrsim \log(n)$, this factor satisfies $C(n,d_e) = \mathcal{O}(1)$, so that the latter aspect ratio scales as $\frac{d_e \log(d_e)}{m}$.

Our proof generalizes the results and analysis techniques from the work of J.~Tropp~\cite{tropp2011improved} who treated the specific case $D=I_d$, and it relies on two powerful matrix inequalities, namely, Lieb's and the matrix Bernstein inequalities~\cite{tropp2015introduction, vershynin2018high}. We defer it to Appendix~\ref{SectionProofSRHTMatrices}. We note that similar concentration bounds were obtained by~\cite{cohen2016optimal} using different analysis techniques.
\btheos 
\label{TheoremMainBoundSRHT} 
Let $\rho \in (0,1)$ and $m \gre C(n,d_e) \cdot \frac{d_e \log(d_e)}{\rho}$. Then it holds with probability at least $1-9/d_e$ that $\lambda_\rho \less \gamma_d \less \gamma_1 \less \Lambda_\rho$ where $\lambda_\rho \defn 1-\|D\|_2^2 \sqrt{\rho}$ and $\Lambda_\rho \defn 1+\|D\|_2^2 \sqrt{\rho}$.
\etheos 
As already discussed in the previous section, the operator norm $\|D\|_2$ might be unknown in practice, but one can replace $\|D\|_2$ by $1$ as follows.
\bde[Practical parameters for the SRHT]
\label{DefTargetsSRHT}
Given $\rho \in (0,1)$, we define the bounds $\lambda_\rho \defn 1-\sqrt{\rho}$ and $\Lambda_\rho \defn 1+\sqrt{\rho}$. We denote the corresponding algorithmic parameters by $\mu_\text{gd}(\rho) \defn \mu_\text{gd}(\lambda_{\rho}, \Lambda_{\rho})$, $\mu_\text{p}(\rho) \defn \mu_\text{p}(\lambda_{\rho}, \Lambda_{\rho})$ and $\beta_\text{p}(\rho) \defn \beta_\text{p}(\lambda_{\rho}, \Lambda_{\rho})$, and the corresponding convergence rates $c_\text{gd}(\rho) \defn c_{\text{gd}}( \lambda_{\rho},\Lambda_{\rho})$ and $c_\text{p}(\rho) \defn c_{\text{p}}(\lambda_{\rho},\Lambda_{\rho})$.
\ede

\section{An adaptive method free of the knowledge of the effective dimension}
\label{SectionAdaptiveMethod}

We propose a novel adaptive method with time-varying sketch size. Our algorithm does not require the knowledge of $d_e$, but still achieves a fast rate of convergence while keeping $m \lesssim d_e$. 

Our method is based on monitoring \emph{an approximation of} the improvement ratio $C_t \defn \frac{\delta_{t+1}}{\delta_t}$. Given a threshold $\overline C$, it proceeds as follows. Starting from an arbitrary initial sketch size (say $m=1$), we compute at time $t$ a gradient-IHS update $x_{t+1}$. If $C_t \lesssim \overline C$, then we accept the update $x_{t+1}$. Otherwise, we reject the update $x_{t+1}$, increase the sketch size by a constant factor (say $m \leftarrow 2m$) and re-compute the sketched matrix $SA$. Since only updates with sufficient improvement are accepted, this method achieves a convergence rate smaller than the chosen threshold $\overline C$. Importantly, with, for instance, Gaussian embeddings, according to Theorems~\ref{TheoremDeterministicRateGradientMethod} and~\ref{TheoremGaussianConcentrationEllipsoids}, as soon as the sketch size becomes larger than $\Omega(d_e / \overline{C})$ then all the updates are accepted, so that the number of rejected updates $K$ is finite with $K \lesssim \log(d_e / \overline C) / \log(2)$. However, computing the exact improvement ratio $C_t$ requires the knowledge of $\overline A x^*$, and we alleviate this difficulty as described next.

We provide a proxy of the improvement ratio which is especially compatible with the Gradient- and Polyak-IHS updates. We introduce the \emph{approximate} error $r_t \defn \frac{1}{2} \|C_S^{-\frac{1}{2}} \overline U^\top \overline A (x_t - x^*)\|^2$, and the \emph{approximate} ratio $c_t \defn \frac{r_{t+1}}{r_t}$. In the next result, we relate the approximate error vector $r_t$ with a quantity that can be efficiently computed. We defer the proof to Appendix~\ref{prooflemmanewtondecrement}.
\blems[Sketched Newton decrement]
\label{lemmanewtondecrement}
It holds that $r_t = \frac{1}{2} g_t^\top H_S^{-1} g_t$, where $g_t \defn \nabla f(x_t)$.
\elems
Since the IHS forms at each iteration the descent direction $H_S^{-1}g_t$, it is fast to additionally compute the sketched Newton decrement\footnote{In the optimization literature~\cite{boyd2004convex}, the Newton decrement at $x$ of a twice differentiable, convex function $f$ is defined as $\frac{1}{2}\nabla f(x)^\top \nabla^2 f(x)^{-1} \nabla f(x)$.} $r_t = \frac{1}{2} g_t^\top H_S^{-1} g_t$ and the approximate improvement ratio $c_t = r_{t+1} / r_t$. Consequently, we can efficiently monitor the ratio $c_t$ as opposed to $C_t$ in order to adapt the sketch size. Provided that $c_t$ and $C_t$ are close enough, this would yield the desired performance. We describe our proposed method in Algorithm~\ref{AlgorithmAdaptive}. 

\begin{algorithm}
	\label{AlgorithmAdaptive}
	\DontPrintSemicolon
	\SetKwInOut{Input}{Input}
	\Input{$A \in \mathbb{R}^{n \times d}$, $b \in \mathbb{R}^n$, $\nu > 0$, initial sketch size $m \gre 1$, initial points $x_0, x_1 \in \real^d$, target convergence rates $\overline c_\text{gd}, \overline c_\text{p} \in (0,1)$, gradient descent step size $\mu_\text{gd}$, Polyak step size $\mu_\text{p} \gre 0$ and momentum parameter $\beta_\text{p} \gre 0$}
	Sample $S \in \real^{m \times n}$ and compute $S_A = SA$.\\
	Compute $g_1 = \nabla f(x_1)$, $\widetilde g_1 = H_S^{-1}g_1$ and $r_1 = \frac{1}{2} g_1^\top \widetilde g_1$.\\
	\For{$t= 1, 2,\dots,T-1$}{
		Compute $x_\text{p}^+ = x_t - \mu_\text{p}\, \widetilde g_t + \beta_\text{p} (x_t \!-\! x_{t-1})$, $g^+_p = \nabla f(x^+_\text{p})$, $\widetilde{g}^+_\text{p} = H_S^{-1} g^+_p$, $r^+_p = \frac{1}{2}{g^+_\text{p}}^\top \widetilde{g}^+_p$.\\
		Compute the Polyak-IHS improvement ratio $c^+_\text{p} = \left(\frac{r^+_\text{p}}{r_1}\right)^{\frac{1}{t}}$.\\
		\eIf{$c^+_\text{p} \less \overline c_\text{p}$}{
		    Set $x_{t+1} = x^+_\text{p}$, $g_{t+1} = g^+_\text{p}$, $\widetilde{g}_{t+1} = \widetilde{g}^+_\text{p}$ and $r_{t+1} = r^+_\text{p}$.
		}{
		Compute $x^+_\text{gd} = x_t - \mu_\text{gd} \, \widetilde g_t$, $g^+_\text{gd} = \nabla f(x^+_\text{gd})$, $\widetilde{g}^+_\text{gd} = H_S^{-1} g^+_\text{gd}$ and $r^+_{\text{gd}} = \frac{1}{2}{g^+_{\text{gd}}}^\top \widetilde{g}^+_{\text{gd}}$.\\
		Compute the gradient-IHS improvement ratio $c^+_\text{gd} = \frac{r_\text{gd}^+}{r_t}$.\\
		\eIf{$c^+_\text{gd} \less \overline c_\text{gd}$}{
		        Set $x_{t+1} = x^+_\text{gd}$, $g_{t+1} = g^+_\text{gd}$, $\widetilde{g}_{t+1} = \widetilde{g}^+_\text{gd}$ and $r_{t+1} = r_\text{gd}^+$.
		    }{
		        Set $m \defn 2m$, sample $S \in \real^{m \times n}$ and compute $S_A = S\cdot A$.\\
		        Set $\widetilde{g}_t \defn H_S^{-1} g_t$ and return to Step 4.
		    }
		}
	}
	Return $x_T$.
	\caption{Adaptive Polyak-IHS method.}
\end{algorithm}
Note that Algorithm~\ref{AlgorithmAdaptive} computes first a Polyak-IHS update. According to Theorem~\ref{TheoremDeterministicRatePolyak}, the relative error of the Polyak-IHS update cannot be tightly controlled in finite-time, but only asymptotically as $t \to +\infty$. This makes difficult to provide guarantees using only the Polyak-IHS update based on monitoring an approximate improvement ratio. Therefore, if the Polyak-IHS update fails, Algorithm~\ref{AlgorithmAdaptive} computes a gradient-IHS update, whose improvement between two successive iterates can be tightly controlled according to Theorem~\ref{TheoremDeterministicRateGradientMethod}. Hence, Algorithm~\ref{AlgorithmAdaptive} may compute both updates in order to benefit either from the acceleration of the latter or from the hard convergence guarantees of the former. If both updates do not make enough progress then the sketch size is increased.

\subsection{Convergence guarantees}

We now state high-probability guarantees on the performance of Algorithm~\ref{AlgorithmAdaptive}. We show that the sketch size and the number of rejected steps remain bounded, i.e., $m = \mathcal{O}(d_e / \rho)$ and $K = \mathcal{O}(\log(d_e / \rho))$ for Gaussian embeddings, whereas $m = \mathcal{O}(d_e \log(d_e) / \rho)$ and $K = \mathcal{O}(\log(d_e \log(d_e) / \rho))$ for the SRHT. Further, the convergence rate roughly scales as $\rho^t$. We defer the proofs of the next two results to Appendices~\ref{prooftheoremadaptivesketchsizegaussian} and~\ref{prooftheoremadaptivesketchsizesrht}.
\btheos[Gaussian embeddings]
\label{theoremadaptivesketchsizegaussian}
Let $\rho \less 0.18$ and $\eta \less 0.01$. Suppose that we run Algorithm~\ref{AlgorithmAdaptive} with $\overline c_{\text{gd}} = c_{\text{gd}}(\rho,\eta)$, $\overline c_{\text{p}} = c_{\text{p}}(\rho,\eta)$, $\mu_\text{gd}=\mu_\text{gd}(\rho,\eta)$, $\mu_p=\mu_p(\rho,\eta)$ and $\beta_p=\beta_p(\rho,\eta)$ (see Definition~\ref{DefTargetsGaussian}), and, with an initial sketch size $m_\text{initial} \gre 1$. Then, it holds with probability at least $1-8 e^{-d_e \eta / 2}$ that, across all iterations, the sketch size remains bounded as
\begin{align}
\label{eqnsketchsizegaussian}
    m \less 2\,c_0 \cdot \frac{d_e}{\rho}\,,
\end{align}
where $c_0$ is a numerical constant which satisfies $c_0 \less 5$. Further, the number of rejected updates is upper bounded as
\begin{align}
\label{eqnrejectedstepsgaussian}
    K \less \frac{\log\left(\frac{c_0 \,d_e}{m_\text{initial}\rho}\right)}{\log 2} + 1\,.
\end{align}
Moreover, at any fixed iteration $t \gre 1$, it holds with probability at least $1-8e^{-d_e \eta/2}$ that the relative error satisfies
\begin{align}
\label{eqncvrategaussian}
    \frac{\delta_t}{\delta_1} \less 9 \, \left(1+\frac{\sigma_1^2}{\nu^2}\right) \max\left\{1,\frac{d_e}{m_\text{initial}}\right\} c_\text{gd}(\rho,\eta)^{t-1} \,.
\end{align}
\etheos 
\btheos[SRHT]
\label{theoremadaptivesketchsizesrht}
Fix $\rho \in (0,1)$. Suppose that we run Algorithm~\ref{AlgorithmAdaptive} with $\overline c_{\text{gd}} = c_{\text{gd}}(\rho)$, $\overline c_{\text{p}} = c_{\text{p}}(\rho)$, $\mu_\text{gd}=\mu_\text{gd}(\rho)$, $\mu_\text{p}=\mu_\text{p}(\rho)$ and $\beta_\text{p}=\beta_\text{p}(\rho)$ (see Definition~\ref{DefTargetsSRHT}), and, with an initial sketch size $m_\text{initial} \gre 1$. Denote $a_{\rho} \defn \frac{1+\sqrt{\rho}}{1-\sqrt{\rho}}$. Then, it holds with probability at least $1-\frac{9}{d_e}$ that, across all iterations, the sketch size remains bounded as
\begin{align}
\label{eqnsketchsizesrht}
    m \less 2 \,a_{\rho} C(n,d_e) \frac{d_e \log d_e}{\rho}\,,
\end{align}
and the number of rejected updates is upper bounded as
\begin{align}
\label{eqnrejectedstepssrht}
    K \less \frac{\log\left(a_{\rho} C(n,d_e) \frac{d_e \log d_e}{m_\text{initial}\rho}\right)}{\log 2} + 1\,.
\end{align}
Moreover, it holds almost surely that, across all iterations, the relative error satisfies
\begin{align}
\label{eqncvratesrht}
    \frac{\delta_t}{\delta_1} \less 2 \, \left(1+\frac{\sigma_1^2}{\nu^2}\right) c_\text{gd}(\rho)^{t-1}\,.
\end{align}
\etheos 
The bound~\eqref{eqnsketchsizesrht} on the sketch size is weaker with the SRHT, which requires an additional factor $\log d_e$. This logarithmic oversampling factor was shown to be necessary for other concentration bounds (see, for instance, the discussions in~\cite{halko2011finding, tropp2011improved}). On the other hand, the bound~\eqref{eqncvrategaussian} on the relative error has an additional factor $\frac{d_e}{m_\text{initial}}$ when $m_\text{initial} \less d_e$ with Gaussian embeddings. According to our proof of Theorem~\ref{theoremadaptivesketchsizesrht}, this follows from the orthogonality of the SRHT which causes less distortions than an i.i.d.~Gaussian embedding, especially when the embedding dimension is small.

\subsection{Time and space complexity}

We consider here the SRHT for which computing $SA$ is faster than Gaussian projections. We have the following complexity result, whose proof is deferred to Appendix~\ref{ProofTheoremTotalComplexity}.
\btheos
\label{TheoremTotalComplexity}
Let $\varepsilon \in (0,1/2)$ be a given precision such that $\varepsilon \less \frac{\nu^2}{\nu^2 + \sigma_1^2}$. Under the hypotheses of Theorem~\ref{theoremadaptivesketchsizesrht}, it holds with probability at least $1-\frac{9}{d_e}$ that the number of iterations to reach a solution $x_T$ such that $\delta_T / \delta_1 \less \varepsilon$ satisfies $T = \mathcal{O}(\frac{\log(1/\varepsilon)}{\log(1/\rho)})$. Thus the total time complexity $\mathcal{C}_\varepsilon$ of Algorithm~\ref{AlgorithmAdaptive} verifies
\begin{align*}
	\mathcal{C}_\varepsilon = \mathcal{O}\left(\log\!\left(d_e/\rho\right) (nd \log\!\left(d_e/\rho\right) + \, \frac{d_e^2 \log^2 \!d_e}{\rho^2} \,d ) + nd \,\frac{\log(1/\varepsilon)}{\log(1/\rho)} \right)\,.
\end{align*}
\etheos 
The time complexity $\mathcal{C}_\varepsilon$ is decomposed into three terms. Sketching the data matrix takes $\mathcal{O}(nd \log(d_e/\rho))$ time. The cost $\mathcal{O}(\frac{d_e^2 \log^2\!d_e}{\rho^2} \, d)$ corresponds to computing a factorization of $H_S$ using the Woodbury identity (see Appendix~\ref{ProofTheoremTotalComplexity} for details). These two costs are multiplied by an extra factor $\mathcal{O}(\log(d_e/\rho))$ which is the maximum number of rejected steps. The last term is the per-iteration complexity $\mathcal{O}(nd)$ times the number of iterations $T = \mathcal{O}(\frac{\log(1/\varepsilon)}{\log(1/\rho)})$. In contrast, other state-of-the-art randomized preconditioning methods~\cite{rokhlin2008fast, avron2010blendenpik, meng2014lsrn} prescribe the sketch size $m =\frac{d \log d}{\rho}$ and they are also decomposed into three steps: sketching, factoring, and iterating. Sketching with the SRHT also costs $\mathcal{O}(nd \log(d/\rho))$ and the factoring step takes $\mathcal{O}(\frac{d^3 \log^2 \!d}{\rho^2})$ time. The iteration part costs $\mathcal{O}(nd\,\frac{\log(1/\varepsilon)}{\log(1/\rho)})$. This yields the total complexity $\mathcal{C}_\text{other} = \mathcal{O}(nd \log(d/\rho) + \frac{d^3\log^2 \!d}{\rho^2} + nd \, \frac{\log(1/\varepsilon)}{\log(1/\rho)})$. Thus, even with the extra factor $\log(d_e/\rho)$ due to the rejected steps, our adaptive method improves on the sketching plus factor costs especially when the effective dimension $d_e$ is much smaller than the data dimension $d$ and thus, on the total complexity.

Regarding space complexity, our method requires $\mathcal{O}(d \cdot d_e \log d_e/\rho)$ space to store the sketched matrix $SA$ whereas the other preconditioning methods needs $\mathcal{O}(d^2 \log d / \rho)$. This is a significant improvement when $d_e$ is much smaller than $d$.

\begin{remark}
Our results developed so far are relevant for a dense data matrix $A$. On the other hand, it is also of great practical interest to develop efficient methods which address the case of sparse data matrices. If the data matrix $A$ has a few non-zero entries, then embeddings for which the computational complexity of forming $SA$ scales as $\mathcal{O}(\text{nnz}(A))$ may be more relevant for our adaptive method. Many deviation bounds similar to those we present in Theorems~\ref{TheoremGaussianConcentrationEllipsoids} and~\ref{TheoremMainBoundSRHT} exist for sparse embeddings (see, for instance,~\cite{chowdhury2018iterative, holodnak2015randomized, cohen2016optimal}). We leave the analysis of our adaptive method with sparse embeddings to future work.
\end{remark}

%\subsection{other ideas}
%Using uniform row sampling, it's known that we get subspace approximation when $m\gtrsim \mu \log d$ where $\mu $ is the maximum leverage score of $A$. Therefore, the adaptive algorithm with uniform sampling sketch provably converges without the knowledge of $\mu$. If the sketch size increases beyond $O(d)$ during iterations, we can switch to another sketching strategy. We can extend the adaptive method to Newton sketch, where we can apply it to arbitrary functions. We can approximately perform the operation $(A^TS^TSA)^{-1}g$ using conjugate gradient and derive an approximation tolerant bound refreshing vs not refreshing. An interesting thing is that the refreshed sketch converges with the optimal rate \emph{non-asymptotically} with no momentum.

\section{Numerical experiments}

We carry out numerical simulations of Algorithm~\ref{AlgorithmAdaptive} and we compare it to standard iterative solvers, that is, the CG method and the randomized preconditioned CG (pCG)~\cite{rokhlin2008fast}. Numerical simulations were carried out on a 512Gb desktop station and implemented in Python using its standard numerical linear algebra modules\footnote{Code is publicly available at https://github.com/jonathanlctt/eff\_dim\_solver}. 

We consider two evaluation criteria: (i) the cumulative time to compute the solutions up to a given precision $\varepsilon > 0$ along an entire regularization path (several values of $\nu$ in decreasing order) and the memory space required by each sketching-based algorithm as measured by the sketch size $m$, and, (ii) the same criteria but for a fixed value of $\nu > 0$.

We present in Figures~\ref{FigCIFARMNISTPath} and~\ref{FigCIFARMNISTnu} results for two standard datasets (see Appendix~\ref{sectionadditionalnumericalexperiments} for additional experiments): (i) one-vs-all classification of MNIST digits and (ii) one-vs-all classification of CIFAR10 images.

Except for very large values of the regularization parameter $\nu > 0$ for which the regularized least-squares problem~\eqref{EqnMain} is well-conditioned so that the conjugate gradient method is very efficient, we observe that our method is the fastest and requires less memory space than pCG for computing both the solutions of the entire regularization path and for a fixed value of $\nu$. In particular, pCG uses $m = \frac{d}{\rho}$ for Gaussian embeddings and $m=\frac{d \log d}{\rho}$ for the SRHT. Note that, without a priori knowledge or estimation of the effective dimension $d_e$, these are the best statistical lower bounds on the sketch size known for pCG in order to guarantee convergence. Thus pCG is especially slower at the beginning because the factorization cost scales as $\mathcal{O}(d^3)$ and it requires memory space  $\mathcal{O}(d^2)$. In contrast, our method starts with $m=1$ and $m$ does not exceed $\mathcal{O}(d_e / \rho)$ for Gaussian embeddings and $\mathcal{O}(d_e \log d_e / \rho)$ for the SRHT, as predicted by Theorems~\ref{theoremadaptivesketchsizegaussian} and~\ref{theoremadaptivesketchsizesrht}. Our adaptive sketch size remains sometimes much smaller than these theoretical upper bounds, and we still have a fast rate of convergence. 

We observe in practice that, in Algorithm~\ref{AlgorithmAdaptive}, the Polyak-IHS update is often rejected compared to the gradient-IHS update, especially with the SRHT. Therefore, in addition to Algorithm~\ref{AlgorithmAdaptive}, we consider a variant which does not compute the Polyak-IHS update but only the gradient-IHS update. This variant enjoys exactly the same convergence guarantees as presented in Theorems~\ref{theoremadaptivesketchsizegaussian} and~\ref{theoremadaptivesketchsizesrht}. Since it computes only a single candidate update, this variant is faster than Algorithm~\ref{AlgorithmAdaptive} in the case where the Polyak-IHS update is often rejected.

\begin{figure}[h!]
	\centering
	\includegraphics[width=0.8\linewidth]{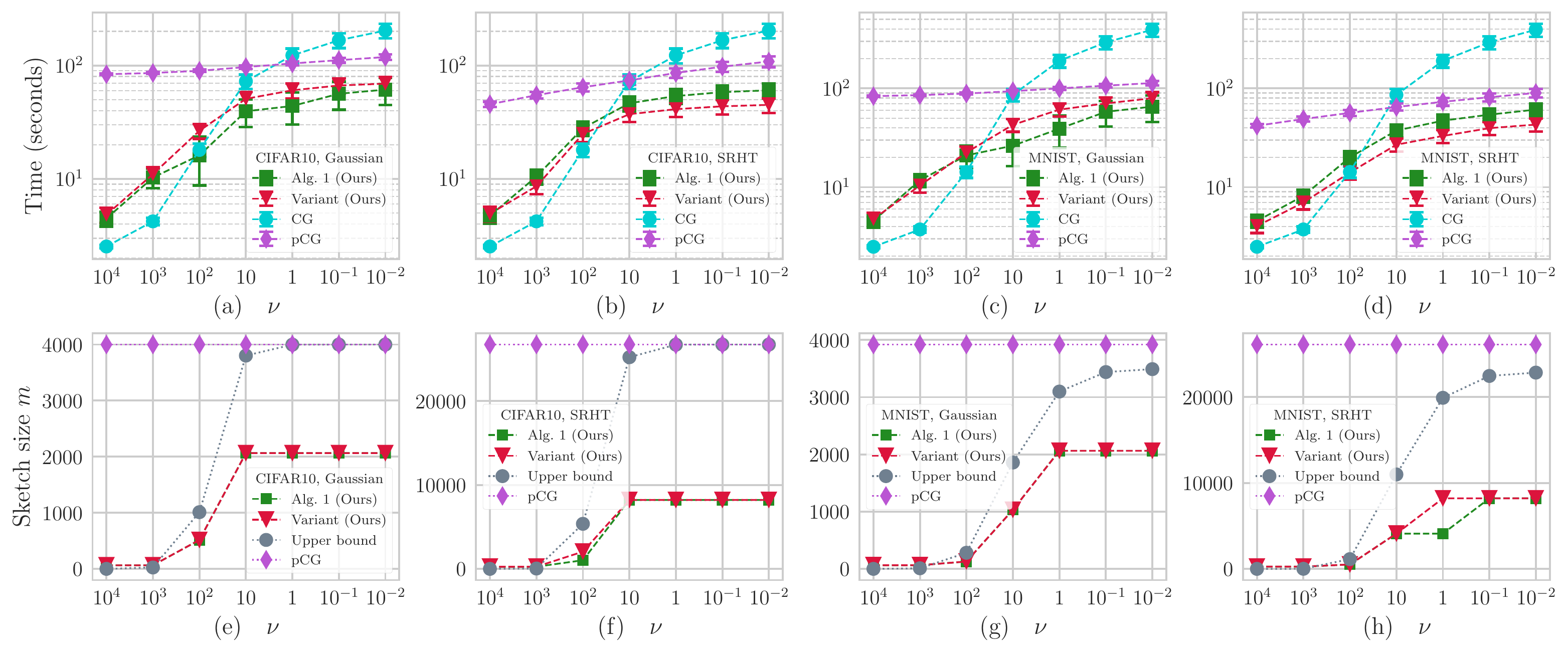}
	\caption{CIFAR10 and MNIST datasets: comparison of CG, pCG, Algorithm~\ref{AlgorithmAdaptive} and a variant of Algorithm~\ref{AlgorithmAdaptive} which only computes gradient-IHS updates. We consider an entire regularization path $\nu \in \{10^j \mid j=4,\dots,-2\}$. For each algorithm, we start with the largest value $\nu = 10^4$. For $j \less 3$, we initialize each algorithm at the previous solution $\widetilde x$ found for $j+1$. For each value of $\nu$, we stop the algorithm once $\varepsilon=10^{-10}$-precision is reached. Each run is averaged over $30$ independent trials. Mean standard deviations are reported in the form of error bars.}
	\label{FigCIFARMNISTPath}
\end{figure}

\begin{figure}[h!]
	\centering
	\includegraphics[width=0.8\linewidth]{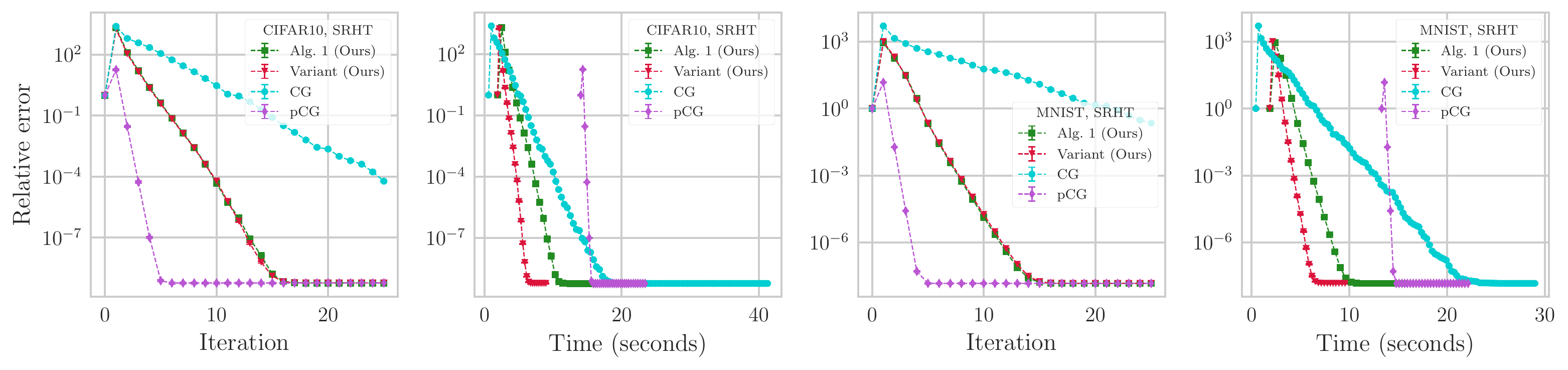}
	\caption{CIFAR10 and MNIST datasets: comparison of CG, pCG, Algorithm~\ref{AlgorithmAdaptive} and our variant of Algorithm~\ref{AlgorithmAdaptive} using gradient-IHS updates only. We fix the value of the regularization parameter $\nu=10$. Each run is averaged over $30$ independent trials.}
	\label{FigCIFARMNISTnu}
\end{figure}

\section*{Broader Impact}

We believe that the proposed method in this work can have positive societal impacts. Our algorithm can be applied in massive scale distributed learning and optimization problems encountered in real-life problems. The computational effort can be significantly lowered as a result of adaptive dimension reduction. Consequently energy costs for optimization can be significantly reduced.

\begin{ack}
This work was partially supported by the National Science Foundation under grants IIS-1838179 and ECCS-2037304, Facebook Research, Adobe Research and Stanford SystemX Alliance.
\end{ack}

%% BIBLIOGRAPHY 

%\bibliography{effdim}

%\input{effdim}

\newpage

\appendix

\section{Additional results}
\label{sectionadditionalresults}

\subsection{Numerical experiments with synthetic datasets}
\label{sectionadditionalnumericalexperiments}

\begin{figure}[h!]
	\centering
	\includegraphics[width=0.8\linewidth]{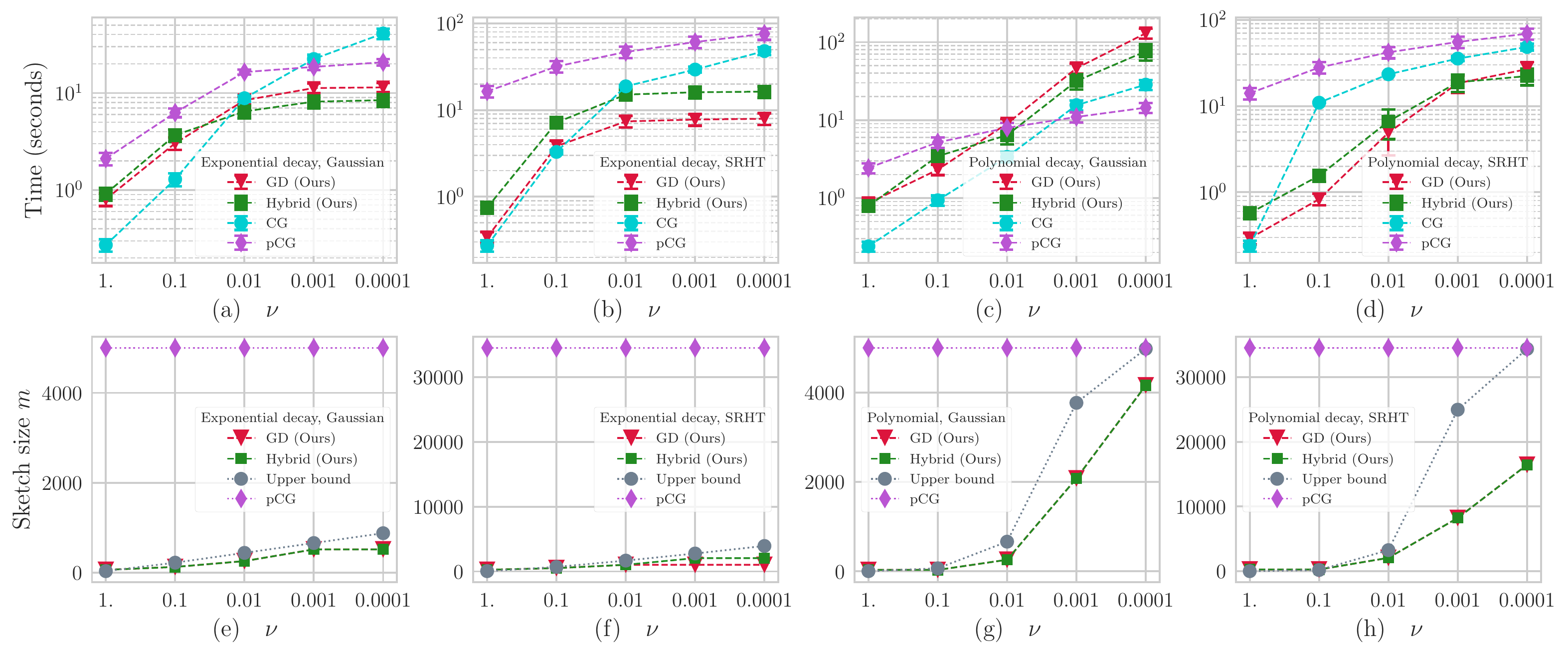}
	\caption{Exponential and polynomial spectral decays: comparison of CG, pCG, Algorithm~\ref{AlgorithmAdaptive} and a variant of Algorithm~\ref{AlgorithmAdaptive} which only computes gradient-IHS updates. We consider an entire regularization path $\nu \in \{10^j \mid j=0,\dots,-4\}$. For each algorithm, we start with the largest value $\nu = 1$. For $j \less 3$, we initialize each algorithm at the previous solution $\widetilde x$ found for $j+1$. For each value of $\nu$, we stop the algorithm once $\varepsilon=10^{-10}$-precision is met. We observe that pCG is slow at the beginning due to forming and factoring the $m \times d$ sketched matrix $S\cdot A$ with $m\approx d$. In contrast, our methods start with $m=1$ and the varying sketch size remains much smaller than that of pCG. This leads to better time and memory space performance, except for the case of Gaussian embeddings and polynomial decays. In the latter case, our method is slowed down by Gaussian projections which are expensive. But with the SRHT, our method has the best performance. Each run is averaged over $30$ independent trials. Mean standard deviations are reported in the form of error bars. }
	\label{FigSyntheticPath}
\end{figure}

Here, we consider a synthetic dataset with $A$ having exponential spectral decay $\sigma_j = 0.95^j$ for $j=1,\dots,d$. The observation vector is generated as follows, $b=Ax_\text{pl} + \eta$, where $x_\text{pl}$ is a planted vector with $\frac{1}{\sqrt{d}}\mathcal{N}(0,1)$ independent entries and $\eta$ is a vector of Gaussian noise $\frac{1}{\sqrt{n}}\mathcal{N}(0,I_n)$. We also consider the similar synthetic dataset but with polynomially decaying singular values $\sigma_j = 1/j$ for $j=1,\dots,d$. Results are reported in Figure~\ref{FigSyntheticPath}.

\subsection{The underdetermined case $n \less d$}
\label{SectionUnderdetermined}

A dual of the problem~\eqref{EqnMain} is
\begin{align*}
    z^* \defn \argmin_{z \in \real^n} \left\{\frac{1}{2} \|A^\top z\|^2 + \frac{\nu^2}{2} \|z\|^2 - b^\top z\right\}\,,
\end{align*}
and one can map the optimal dual solution $z^*$ to the primal one using the relationship
\begin{align}
\label{EqnDualPrimalMap}
    x^* = A^\top z^*\,.
\end{align}
The dual problem fits into the primal overdetermined framework we consider in the main body of this manuscript. Indeed, we have that
\begin{align}
\label{EqnDualMain}
    z^* = \argmin_{z \in \real^n} \left\{g(z) \defn \frac{1}{2} \|A^\top z - \widehat b\|^2 + \frac{\nu^2}{2} \|z\|^2 \right\}\,,
\end{align}
where $\widehat b = A^\dagger b$ and $A^\dagger$ is the pseudo-inverse of $A$. One might wonder whether $\widehat b$ needs to be computed in order to apply the previous framework to the dual overdetermined case: this is not the case. Indeed, in Algorithm~\ref{AlgorithmAdaptive}, the observation vector $b$ only appears in the gradient formula, as $\nabla f(x_t) = A^\top (Ax_t - b)$. For the dual problem~\eqref{EqnDualMain}, we have
\begin{align*}
    \nabla g(z_t) = A (A^\top z_t - \widehat{b}) = A A^\top z_t - b\,.
\end{align*}
That is, the gradient is easily computed and Algorithm~\ref{AlgorithmAdaptive} can be applied to the dual problem~\eqref{EqnDualMain} with the exact same guarantees for the sketch size and the number of rejected steps as in Theorems~\ref{theoremadaptivesketchsizegaussian} and~\ref{theoremadaptivesketchsizesrht}, while having guarantees on the error
\begin{align*}
    \varepsilon_t \defn \frac{1}{2} \|A^\top (z_t - z^*)\|^2 + \frac{\nu^2}{2} \|z_t - z^*\|^2\,,
\end{align*}
Using the map $x_t = A^\top z_t$, the notation $\delta_t = \frac{1}{2}\|A(x_t - x^*)\|^2 + \frac{\nu^2}{2}\|x_t-x^*\|^2$ and assuming that $z_0 = 0$ so that $\varepsilon_0 = f(x^*) / \nu^2$, we obtain with Algorithm~\ref{AlgorithmAdaptive} that $\varepsilon_t \lesssim \rho^t \varepsilon_0$, and consequently
\begin{align*}
    \frac{1}{2} \|A(x_t - x^*)\|^2 + \frac{\nu^2}{2} \|x_t-x^*\|^2 &= \frac{1}{2} \|AA^\top (z_t - z^*)\|^2 + \frac{\nu^2}{2} \|A^\top (z_t-z^*)\|^2\\
    & \less \sigma_1(A)^2 \cdot \varepsilon_t \\
    & \less \frac{\sigma_1(A)^2 f(x^*)}{\nu^2} \cdot \rho^t\,.
\end{align*}
Thus, the total number of iterations to reach $\varepsilon$-relative accuracy for $x_t$ becomes
\begin{align*}
    T = \mathcal{O}\left(\frac{\log(1/\varepsilon) + \log(\sigma_1(A)^2 / \nu^2) + \log(f(x^*)/\delta_0))}{\log(1/\rho)}\right)\,.
\end{align*}
Under the hypothesis $\varepsilon \less \frac{\nu^2}{\nu^2 + \sigma_1(A)^2}$ of Theorem~\ref{TheoremTotalComplexity} and the additional hypothesis $\frac{f(x^*)}{\delta_0} \less \varepsilon^{-1}$, this number of iterations scales as 
\begin{align*}
    T = \mathcal{O}\left(\log(1/\varepsilon) / \log(1/\rho)\right)\,.
\end{align*}
Consequently, we obtain the same total computational complexity (both in time and space) as stated in Theorem~\ref{TheoremTotalComplexity} to reach an approximate solution $x_t$ with $\varepsilon$-relative accuracy.

\section{Proof of main results}

\subsection{Proof of Theorems~\ref{TheoremDeterministicRateGradientMethod} and~\ref{TheoremDeterministicRatePolyak}}
\label{ProofTheoremDeterministicRates}

We denote by $A = U \Sigma V^\top$ a singular value decomposition of the matrix $A$, where $U = [u_1, \hdots, u_d] \in \real^{n \times d}$ has orthonormal columns, $V = [v_1, \hdots,v_d] \in \real^{d \times d}$ has orthonormal columns, and $\Sigma = \text{diag}(\sigma_1, \hdots, \sigma_d)$, with $\sigma_1 \gre \hdots \gre \sigma_d > 0$. 

We denote $D = \textrm{diag}\left(\frac{\sigma_1}{\sqrt{\sigma_1^2+ \nu^2}}, \dots, \frac{\sigma_d}{\sqrt{\sigma_d^2+ \nu^2}}\right)$, $D^\prime=\textrm{diag}\left(\frac{\nu}{\sqrt{\sigma_1^2 + \nu^2}}, \dots, \frac{\nu}{\sqrt{\sigma_d^2 + \nu^2 }}\right)$, and further,
\begin{align*}
    \bar{U} \defn \begin{bmatrix} U D \\ V D^\prime\end{bmatrix}\,,\qquad \bar{\Sigma} \defn \textrm{diag}\left(\sqrt{\sigma_1^2 + \nu^2}, \dots, \sqrt{\sigma_d^2 + \nu^2}\right)\,.
\end{align*}
Note that $\bar{A} = \bar{U} \bar{\Sigma} V^\top$. Indeed,
\begin{align*}
    \bar{U} \bar{\Sigma} V^\top = \begin{bmatrix} UD \bar{\Sigma} V^\top \\ V D^\prime \bar{\Sigma} V^\top \end{bmatrix} = \begin{bmatrix} U \Sigma V^\top \\ V (\nu \cdot I_d) V^\top \end{bmatrix} = \begin{bmatrix} A \\ \nu \cdot I_d \end{bmatrix}\,.
\end{align*}
Further, the columns of $\bar{U}$ are orthonormal, and the matrix $\bar{\Sigma}$ is diagonal with non-negative entries, so that $\bar{U} \bar{\Sigma} V^\top$ is a singular value decomposition of $\bar{A}$. 

Given an embedding $S \in \real^{m \times n}$, denote by $\bar{S}$ the $(m+d) \times (n+d)$ block-diagonal matrix $\begin{bmatrix} S & 0 \\ 0 & I_d\end{bmatrix}$. Denote $\bar b = \begin{bmatrix} b\\0 \end{bmatrix}$. We have that $\bar{A}^\top \bar{S}^\top \bar{S} \bar{A} = A^\top S^\top S A + \nu^2 I_d = H_S$. Consequently, given a step size $\mu \in \real$ and a momentum parameter $\beta \in \real$, the update formula~\eqref{EqnPolyak} of the Polyak-IHS method can be equivalently written as
\begin{align}
\label{EqnPolyakFullStack}
    x_{t+1} = x_t - \mu (\bar{A}^\top \bar{S}^\top \bar{S} \bar{A})^{-1} \bar{A}^\top (\bar{A} x_t - \bar{b}) + \beta (x_t - x_{t-1})\,.
\end{align}
Multiplying the update formula~\eqref{EqnPolyakFullStack} by $\bar{U}^\top \bar{A}$, subtracting $\bar{U}^\top \bar{A} x^*$, using the normal equation $\bar{A}^\top \bar{b} = \bar{A}^\top \bar{A} x^*$ and using the notation $e_t \defn \bar{U}^\top \bar{A}(x_t - x^*)$, we obtain that
\begin{align*}
    e_{t+1} &= e_t -\mu \bar{U}^\top \bar{A}(\bar{A}^\top \bar{S}^\top \bar{S}\bar{A})^{-1}\bar{A}^\top \bar{U} e_t + \beta (e_t - e_{t-1})\\
    &= \left(I - \mu (\bar{U}^\top \bar{S}^\top \bar{S} \bar{U})^{-1} \right) e_t + \beta (e_t - e_{t-1})\,.
\end{align*}
Further, unrolling the expression $\bar{U}^\top \bar{S}^\top \bar{S} \bar{U} = D(U^\top S^\top S U - I_d) D + I_d = C_S$, we find the error recursion
\begin{align}
\label{EqnDynamics}
    \begin{bmatrix} e_{t+1}\\ e_t \end{bmatrix} = \underbrace{\begin{bmatrix} (1+\beta)I_d - \mu C_S^{-1} & -\beta I_d \\ I_d & 0 \end{bmatrix}}_{\defn \, M(\mu,\beta)} \begin{bmatrix} e_t \\ e_{t-1} \end{bmatrix}\,.
\end{align}

\subsubsection{Gradient-IHS method}

For the gradient-IHS method, we have that $\beta = 0$ so that the dynamics~\eqref{EqnDynamics} simplifies to 
\begin{align*}
    e_{t+1} = (I_d - \mu C_S^{-1}) e_t\,.
\end{align*}
Using the fact that $\delta_t = \frac{1}{2} \|e_t\|^2$, we obtain that for any $t \gre 0$,
\begin{align*}
    \frac{\delta_{t+1}}{\delta_t} \less \|I_d - \mu C_S^{-1}\|_2^2\,.
\end{align*}
The eigenvalues of the matrix $I_d - \mu C_S^{-1}$ are given by $1-\frac{\mu}{\gamma_i}$ where the $\gamma_i$'s are the eigenvalues of $C_S$ indexed in non-increasing order. Then, 
\begin{align*}
    \|I_d - \mu C_S^{-1}\|_2 = \max\left\{|1-\frac{\mu}{\gamma_1}|, |1-\frac{\mu}{\gamma_d}|\right\}\,.
\end{align*}
If $\lambda, \Lambda > 0$ are two real numbers such that $\lambda \less \gamma_d \less \gamma_1 \less \Lambda$, then it holds that for any $\mu \gre 0$,
\begin{align*}
    \max\left\{|1-\frac{\mu}{\gamma_1}|, |1-\frac{\mu}{\gamma_d}|\right\} \less \max\left\{|1-\frac{\mu}{\Lambda}|, |1-\frac{\mu}{\lambda}|\right\}\,.
\end{align*}
Picking $\mu = 2 / (\frac{1}{\lambda} + \frac{1}{\Lambda})$ yields that 
\begin{align*}
    \|I_d - \mu C_S^{-1}\|_2 \less \left(\frac{\Lambda - \lambda}{\Lambda + \lambda}\right)\,,
\end{align*}
which is the result claimed in Theorem~\ref{TheoremDeterministicRateGradientMethod}.

\subsubsection{Polyak-IHS method}

Using~\eqref{EqnDynamics} and the fact that $\delta_t = \frac{1}{2}\|e_t\|^2$, we immediately find by recursion that
\begin{align*}
    \left(\frac{\delta_{t+1} + \delta_t}{\delta_1 + \delta_0}\right)^{\frac{1}{t}} \less \|M(\mu, \beta)^t\|_2^\frac{2}{t}\,.
\end{align*}
From Gelfand formula, we obtain that
\begin{align*}
    \limsup_{t \to \infty}\left(\frac{\delta_t}{\delta_0}\right)^{\frac{1}{t}} \less {\rho(M(\mu,\beta))}^2\,,
\end{align*}
where $\rho(M(\mu,\beta))$ is the spectral radius\footnote{The spectral radius of a complex-valued matrix is the largest module of its complex eigenvalues.} of the matrix $M(\mu,\beta)$. Let $C_S = T \Lambda T^\top$ be an eigenvalue decomposition of the positive definite matrix $C_S$ -- where $\Lambda = \textrm{diag}(\gamma_1, \dots, \gamma_d)$ and $\gamma_1 \gre \dots \gamma_d > 0$ --, and define the $(2d) \times (2d)$ permutation matrix $\Pi$ as
\begin{align*}
    \Pi_{i,j} = \begin{cases} 1\,\,\,\textrm{if} \,\,i \,\,\textrm{odd}\,,\,\,j=i\\1\,\,\,\textrm{if} \,\,i \,\,\textrm{even}\,,\,\,j=n+i\\0 \,\,\textrm{otherwise}\end{cases}
\end{align*}
Then, it holds that
\begin{align*}
    \Pi \begin{bmatrix} T & 0 \\ 0 & T \end{bmatrix}^\top M(\mu,\beta) \begin{bmatrix} T & 0 \\ 0 & T \end{bmatrix} \Pi^\top = \begin{bmatrix} M_1(\mu,\beta) & 0 & \dots & 0 \\ 0 & M_2(\mu,\beta) & \dots & 0 \\ \vdots & & \ddots & \vdots \\ 0 & 0 & \dots & M_d(\mu,\beta) \end{bmatrix}
\end{align*}
where $M_i(\mu,\beta) = \begin{bmatrix} 1+\beta - \mu \gamma^{-1}_i & -\beta \\ 1 & 0 \end{bmatrix}$. That is, $M(\mu,\beta)$ is similar to the block diagonal matrix with $2 \times 2$ diagonal blocks $M_i(\mu,\beta)$. To compute the eigenvalues of $M(\mu,\beta)$, it suffices to compute the eigenvalues of all of the $M_i(\mu,\beta)$. For fixed $i$, the eigenvalues of the $2 \times 2$ matrix are roots of the equation $u^2 - (1+\beta - \mu /\gamma_i)u + \beta = 0$. In the case that $1 \gre \beta \gre (1-\sqrt{\mu/\gamma_i})^2$, the roots of the characteristics equations are imaginary, and both have magnitude $\sqrt{\beta}$. Pick $\mu = \mu^* \defn 4 / (1/\sqrt{\Lambda} + 1/\sqrt{\lambda})^2$ and $\beta = \beta^* \defn \left(\frac{\sqrt{\Lambda} - \sqrt{\lambda}}{\sqrt{\Lambda}+\sqrt{\lambda}}\right)^2$, where $\lambda, \Lambda > 0$ are respectively any lower and upper bounds of $\gamma_d$ and $\gamma_1$. Then, we have that $\beta \gre (1-\sqrt{\mu/\gamma_i})^2$ for all $i=1,\dots,d$, so that $\rho(M(\mu,\beta)) \less \sqrt{\beta}$, and this yields the claimed result.
\qed

\subsection{Proof of Theorem~\ref{theoremadaptivesketchsizegaussian}}
\label{prooftheoremadaptivesketchsizegaussian}

We introduce the notation $\overline m = 5\cdot \frac{d_e}{\rho}$.

Either the sketch size always remains smaller than $\overline m$, which is equivalent to 
\begin{align}
    K \less \frac{\log(\overline m / m_\text{initial})}{\log(2)}\,,
\end{align}
in which case the statements~\eqref{eqnsketchsizegaussian} and~\eqref{eqnrejectedstepsgaussian} of Theorem~\ref{theoremadaptivesketchsizegaussian} on the sketch size and the number of rejected steps hold almost surely. 

Otherwise, suppose that for some iteration $t \gre 1$, we have $m > \overline m$. Let $\overline t \gre 1$ be the first such iteration, so that $m \less 2\,\overline m$ and $K \less \frac{\log(\overline m / m_\text{initial})}{\log(2)} + 1$. 

Denote $S$ the sketching matrix sampled at time $\overline t$. Let $\lambda_{\rho/5,\eta}$ and $\Lambda_{\rho/5,\eta}$ be the bounds as given in Definition~\ref{DefTargetsGaussian} (where $\rho$ is replaced by $\rho/5$), and consider the event
\begin{align}
    \mathcal{E}_{\rho/5} \defn \left\{\lambda_{\rho/5,\eta} \less \sigma_\text{min}(C_S) \less \sigma_\text{max}(C_S) \less \Lambda_{\rho/5,\eta}\right\}\,,
\end{align}
which, according to Theorem~\ref{TheoremGaussianConcentrationEllipsoids} and the fact that $m > \overline m$, holds with probability at least $1-8e^{-d_e \eta / 2}$. 

We assume, from now on, that the event $\mathcal{E}_{\rho/5}$ holds. Let $t \gre \overline t$ be any time such that between $\overline t$ and $t$, all updates were accepted (either Polyak- or gradient-IHS), so that the sketch size and sketching matrix are still the same. We claim that \emph{it suffices to prove that the gradient-IHS update at time $t$ is accepted.}

Denote $x_t$ the current iterate, $\delta_t = \frac{1}{2} \|\overline A (x_t - x^*)\|^2$ and $r_t = \frac{1}{2} \|C_S^{-\frac{1}{2}} \overline U^\top \overline A (x_t - x^*)\|^2$. Let $x_\text{gd}^+$ be the gradient-IHS update of Algorithm~\ref{AlgorithmAdaptive}, and denote $\delta^+ \defn \frac{1}{2} \|\overline A(x_\text{gd}^+ - x^*)\|^2$ and $r^+ \defn \frac{1}{2}\|C_S^{-\frac{1}{2}} \overline U^\top \overline A(x_\text{gd}^+ - x^*)\|^2 $. Recall from Lemma~\ref{lemmanewtondecrement} that $r_t$ and $r^+$ are also the sketched Newton decrements at $x_t$ and $x^+$, so that the gradient-IHS improvement ratio computed in Algorithm~\ref{AlgorithmAdaptive} is equal to $\frac{r^+}{r_t}$.

We need the following technical result whose proof is deferred to Appendix~\ref{prooflemmagaussiandetails}.
\blems 
\label{lemmagaussiandetails}
Suppose that $\rho \less 0.18$ and $\eta \less 0.01$. Then, on the event $\mathcal{E}_{\rho/5}$, it holds that
\begin{align}
    \frac{\sigma_\text{max}(C_S)}{\sigma_\text{min}(C_S)} \cdot c_\text{gd}(\rho/5,\eta) \less c_\text{gd}(\rho,\eta)\,.
\end{align}
\elems 
We have that
\begin{align*}
    \frac{\delta^+}{\delta_t} \underset{(i)}{\less} c_\text{gd}(\rho/5,\eta) \underset{(ii)}{\less} \frac{\sigma_\text{min}(C_S)}{\sigma_\text{max}(C_S)}\, c_\text{gd}(\rho,\eta)\,,
\end{align*}
where inequality (i) follows from Theorem~\ref{TheoremDeterministicRateGradientMethod}, and, inequality (ii) from Lemma~\ref{lemmagaussiandetails}. Using $r^+ \less \frac{\delta^+}{\sigma_\text{min}(C_S)}$ and $r_t \gre \frac{\delta_t}{\sigma_\text{max}(C_S)}$, it follows that
\begin{align*}
    \frac{r^+}{r_t} \less \frac{\sigma_\text{max}(C_S)}{\sigma_\text{min}(C_S)} \cdot \frac{\delta^+}{\delta_t} \less c_\text{gd}(\rho,\eta)\,.
\end{align*}
Consequently, the gradient-IHS update $x_\text{gd}^+$ verifies the improvement criterion $\frac{r^+}{r_t} \less c_\text{gd}(\rho,\eta)$, and the update $x_\text{gd}^+$ is not rejected.

In summary, as soon as $m > \overline m$ and provided that $\mathcal{E}_{\rho/5}$ holds, future updates are not rejected. This holds with probability at least $1-8e^{-d_e \eta/2}$, which concludes the proof of the statements~\eqref{eqnsketchsizegaussian} and~\eqref{eqnrejectedstepsgaussian} on the sketch size and the number of rejected steps.

We turn to showing statement~\eqref{eqncvrategaussian}. Fix any iteration $t \gre 1$. By construction of Algorithm~\ref{AlgorithmAdaptive}, it holds almost surely that
\begin{align*}
    \frac{r_t}{r_1} \less \max\{c_\text{gd}(\rho,\eta)^{t-1}, c_\text{p}(\rho,\eta)^{t-1}\} = c_\text{gd}(\rho,\eta)^{t-1}\,.
\end{align*}
Denoting by $S$ the sketching matrix at time $t$, and using that $\delta_t \less \sigma_\text{max}(C_{S}) \cdot r_t$ and $\delta_1 \gre \sigma_\text{min}(C_{S_\text{initial}}) \cdot r_1$, it follows that
\begin{align*}
    \frac{\delta_t}{\delta_1} \less \frac{\sigma_\text{max}(C_{S})}{\sigma_\text{min}(C_{S_\text{initial}})} \cdot \frac{r_t}{r_1} \less \frac{\sigma_\text{max}(C_{S})}{\sigma_\text{min}(C_{S_\text{initial}})} \cdot {c_\text{gd}(\rho,\eta)}^{t-1}\,.
\end{align*}
On the one hand, according to Theorem~\ref{TheoremGaussianConcentrationEllipsoids}, we have that 
\begin{align*}
    \sigma_\text{max}(C_S) &\less \frac{\nu^2}{\sigma_1^2 + \nu^2} + \frac{\sigma_1^2}{\sigma_1^2+\nu^2} \cdot \left(1+\sqrt{(1+3\sqrt{\eta})^2 \frac{d_e}{m_\text{initial}}}\right)^2\,.
\end{align*}
with probability at least $1-8 e^{-\eta d_e / 2}$. Using that $\eta \less 0.01$, $(1+3\sqrt{\eta}) \less 3/2$ and  $(1+\sqrt{\frac{d_e}{m}})^2 \less 4 \, \max\{1, \frac{d_e}{m_\text{initial}}\}$, we obtain
\begin{align*}
    \sigma_\text{max}(C_S) \less 9 \left(\frac{\nu^2}{\sigma_1^2 + \nu^2} + \frac{\sigma_1^2}{\sigma_1^2 + \nu^2} \max\{1, \frac{d_e}{m_\text{initial}}\} \right)\,.
\end{align*}
On the other hand, it holds almost surely that
\begin{align*}
    \sigma_\text{min}(C_{S_\text{initial}}) \gre 1-\|D\|_2^2 = \frac{\nu^2}{\sigma_1^2+\nu^2}\,.
\end{align*}
Combining the latter inequalities, it holds with probability at least $1-8e^{-d_e \eta/2}$ that
\begin{align*}
    \frac{\delta_t}{\delta_1} \less 9 \, \left(1+\frac{\sigma_1^2}{\nu^2}\right) \max\left\{1,\frac{d_e}{m_\text{initial}}\right\} c_\text{gd}(\rho,\eta)^{t-1}\,,
\end{align*}
which concludes the proof.

\subsection{Proof of Theorem~\ref{theoremadaptivesketchsizesrht}}
\label{prooftheoremadaptivesketchsizesrht}

The proof for the SRHT follows steps similar to the Gaussian case. We introduce the notation 
\begin{align}
    \overline m = a_\rho \cdot C(n, d_e) \frac{d_e \log(d_e)}{\rho}\,,
\end{align}
and we recall that $a_\rho \defn \frac{1+\sqrt{\rho}}{1-\sqrt{\rho}}$.

Either the sketch size always remains smaller than $\overline m$. The latter is equivalent to 
\begin{align}
    K \less \frac{\log(\overline m / m_\text{initial})}{\log(2)}\,,
\end{align}
in which case the statements~\eqref{eqnsketchsizesrht} and~\eqref{eqnrejectedstepssrht} of Theorem~\ref{theoremadaptivesketchsizesrht} on the sketch size and the number of rejected steps hold almost surely. 

Otherwise, suppose that for some iteration $t \gre 1$, we have $m > \overline m$. Let $\overline t \gre 1$ be the first such iteration, so that $m \less 2\,\overline m$ and $K \less \frac{\log(\overline m / m_\text{initial})}{\log(2)} + 1$. 

Denote $S$ the sketching matrix sampled at time $\overline t$. Define $\lambda_{\rho/a_\rho} \defn 1-\sqrt{\frac{\rho}{a_\rho}}$ and $\Lambda_{\rho/a_\rho} \defn 1+\sqrt{\frac{\rho}{a_\rho}}$, and consider the event
\begin{align}
    \mathcal{E}_{\rho/a_\rho} \defn \left\{\lambda_{\rho/a_\rho} \less \sigma_\text{min}(C_S) \less \sigma_\text{max}(C_S) \less \Lambda_{\rho/a_\rho}\right\}\,,
\end{align}
which, according to Theorem~\ref{TheoremMainBoundSRHT} and the fact that $m > \overline m$, holds with probability at least $1-\frac{d_e}{9}$. 

We assume, from now on, that the event $\mathcal{E}_{\rho/a_\rho}$ holds. Let $t \gre \overline t$ be any time such that between $\overline t$ and $t$, all updates were accepted (either Polyak- or gradient-IHS), so that the sketch size and sketching matrix are the same. We claim that \emph{it suffices to prove that the gradient-IHS update at time $t$ is accepted.}

Denote $x_t$ the current iterate, $\delta_t = \frac{1}{2} \|\overline A (x_t - x^*)\|^2$ and $r_t = \frac{1}{2} \|C_S^{-\frac{1}{2}} \overline U^\top \overline A (x_t - x^*)\|^2$. Let $x_\text{gd}^+$ be the gradient-IHS update of Algorithm~\ref{AlgorithmAdaptive}, and denote $\delta^+ \defn \frac{1}{2} \|\overline A(x_\text{gd}^+ - x^*)\|^2$ and $r^+ \defn \frac{1}{2}\|C_S^{-\frac{1}{2}} \overline U^\top \overline A(x_\text{gd}^+ - x^*)\|^2 $. Recall from Lemma~\ref{lemmanewtondecrement} that $r_t$ and $r^+$ are also the sketched Newton decrements at $x_t$ and $x^+$, so that the gradient-IHS improvement ratio computed in Algorithm~\ref{AlgorithmAdaptive} is equal to $\frac{r^+}{r_t}$.

We need the following technical result whose proof is deferred to Appendix~\ref{prooflemmasrhtdetails}.
\blems 
\label{lemmasrhtdetails}
On the event $\mathcal{E}_{\rho/a_\rho}$, it holds that $\frac{\sigma_\text{max}(C_S)}{\sigma_\text{min}(C_S)} \less a_\rho$ and $c_\text{gd}(\rho/a_\rho) = \frac{c_\text{gd}(\rho)}{a_\rho}$.
\elems 
We have that
\begin{align*}
    \frac{\delta^+}{\delta_t} \underset{(i)}{\less} c_\text{gd}(\rho/a_\rho) \underset{(ii)}{=} \frac{c_\text{gd}(\rho)}{a_\rho}\,,
\end{align*}
where inequality (i) follows from Theorem~\ref{TheoremDeterministicRateGradientMethod}, and, equality (ii) from the second part of Lemma~\ref{lemmasrhtdetails}. Using $r^+ \less \frac{\delta^+}{\sigma_\text{min}(C_S)}$ and $r_t \gre \frac{\delta_t}{\sigma_\text{max}(C_S)}$, it follows that
\begin{align*}
    \frac{r^+}{r_t} \less \frac{\sigma_\text{max}(C_S)}{\sigma_\text{min}(C_S)} \cdot \frac{\delta^+}{\delta_t} \less \frac{\sigma_\text{max}(C_S)}{\sigma_\text{min}(C_S)} \cdot \frac{c_\text{gd}(\rho)}{a_\rho} &\underset{(i)}{\less} a_\rho \cdot \frac{c_\text{gd}(\rho)}{a_\rho} = c_\text{gd}(\rho)\,,
\end{align*}
where inequality (i) follows from the first part of Lemma~\ref{lemmasrhtdetails}. Consequently, the gradient-IHS update $x_\text{gd}^+$ verifies the improvement criterion $\frac{r^+}{r_t} \less c_\text{gd}(\rho)$, and the update $x_\text{gd}^+$ is not rejected.

In summary, as soon as $m > \overline m$ and provided that $\mathcal{E}_{\rho/a_\rho}$ holds, future updates are not rejected. This holds with probability at least $1-\frac{9}{d_e}$, which concludes the proof of the statements~\eqref{eqnsketchsizesrht} and~\eqref{eqnrejectedstepssrht} on the sketch size and the number of rejected steps.

We turn to showing statement~\eqref{eqncvratesrht}. Fix any iteration $t \gre 1$. By construction of Algorithm~\ref{AlgorithmAdaptive}, it holds almost surely that
\begin{align*}
    \frac{r_t}{r_1} \less \max\{c_\text{gd}(\rho)^{t-1}, c_\text{p}(\rho)^{t-1}\} = c_\text{gd}(\rho)^{t-1}\,.
\end{align*}
Denoting by $S$ the sketching matrix at time $t$, and using that $\delta_t \less \sigma_\text{max}(C_{S}) \cdot r_t$ and $\delta_1 \gre \sigma_\text{min}(C_{S_\text{initial}}) \cdot r_1$, it follows that
\begin{align*}
    \frac{\delta_t}{\delta_1} \less \frac{\sigma_\text{max}(C_{S})}{\sigma_\text{min}(C_{S_\text{initial}})} \cdot \frac{r_t}{r_1} \less \frac{\sigma_\text{max}(C_{S})}{\sigma_\text{min}(C_{S_\text{initial}})} \cdot c_\text{gd}(\rho)^{t-1}\,.
\end{align*}
On the one hand, it holds almost surely that
\begin{align*}
    \sigma_\text{max}(C_S) &= \sup_{\|x\|_2 = 1} \|x\|_2^2 + \langle Dx, (U^\top S^\top S U - I_d) Dx \rangle \\
    & \underset{(i)}{\less} 1 + \sup_{\|x\|_2 \less 1} \langle x, (U^\top S^\top S U - I_d) x \rangle\\
    & \less 1 + \sup_{\|x\|_2 \less 1} \langle x, U^\top S^\top S U x \rangle\\
    & \underset{(ii)}{\less} 2\,,
\end{align*}
where inequality (i) follows from the fact that $\|D\|_2 \less 1$, and inequality (ii) from the fact that $SU$ is a partial orthogonal matrix so that $\|SU\|_2 \less 1$. On the other hand, it holds almost surely that
\begin{align*}
    \sigma_\text{min}(C_{S_\text{initial}}) \gre 1-\|D\|_2^2 = \frac{\nu^2}{\sigma_1^2+\nu^2}\,.
\end{align*}
Combining the latter inequalities, it holds almost surely that
\begin{align*}
    \frac{\delta_t}{\delta_0} \less 2 \, \left(1+\frac{\sigma_1^2}{\nu^2}\right) c_\text{gd}(\rho)^{t-1}\,,
\end{align*}
which concludes the proof.

\qed

\subsection{Proof of Theorem~\ref{TheoremTotalComplexity}}
\label{ProofTheoremTotalComplexity}

According to Theorem~\ref{theoremadaptivesketchsizesrht}, we have with probability at least $1-\frac{9}{d_e}$ that over an entire trajectory, the sketch size and the number of rejected steps satisfy
\begin{align*}
    m = \mathcal{O}(d_e \log d_e / \rho)\,,\quad K = \mathcal{O}(\log(d_e/\rho))\,.
\end{align*}
From now on, we assume that the above event holds.

Then, forming the sketched matrix $SA$ costs at most $\mathcal{O}(nd \log d_e)$ at any iteration. Using the Woodbury matrix identity, the inverse of $H_S$ verifies
\begin{align*}
    H_S^{-1} = \left((SA)^\top SA + \nu^2 I_d\right)^{-1} = \frac{1}{\nu^2}\left( I_d - (SA)^\top (\nu^2 I_{m} +  SA (SA)^\top)^{-1} SA \right)\,.
\end{align*}
To reduce the complexity of solving at each iteration the linear system $H_S \cdot z = \nabla f(x_t)$, one can simply compute and cache a factorization of the matrix $(\nu^2 I_{m} +  SA (SA)^\top)$ which takes time $\mathcal{O}(\frac{d_e^2 \log^2\!d_e}{\rho^2} d)$. Consequently, the total sketching and factor costs scale as $\mathcal{O}(\log (d_e/\rho) \cdot (\frac{d_e^2 \log^2\!d_e}{\rho^2} d + nd \log(d_e / \rho)))$.

The per-iteration cost is that of computing the matrix-vector products $A x_t$ and $A^\top (Ax_t - b)$, which is given by $\mathcal{O}(nd)$. Note that the other main numerical operation consists in solving the linear system $H_S \cdot z = \nabla f(x_t)$. Using the cached factorization of the matrix $(\nu^2 I_{m} +  SA (SA)^\top)$ and the Woodbury identity, this linear system can be solved in time $\mathcal{O}(\frac{d_e \log d_e}{\rho} d)$, which is negligible compared to $\mathcal{O}(nd)$.

According to Theorem~\ref{theoremadaptivesketchsizesrht}, we have almost surely that over an entire trajectory,
\begin{align*}
    \frac{\delta_{t+1}}{\delta_1} \less 2 \cdot (1+\frac{\sigma_1^2}{\nu^2})\cdot {c_\text{gd}(\rho)}^t\,.
\end{align*}
A simple calculation yields that $c_\text{gd}(\rho) = \rho$. Therefore, a sufficient number of iterations $T$ to reach an $\varepsilon$-accurate solution is exactly given by
\begin{align*}
    T = \left\lceil\frac{\log 2 + \log(1+\frac{\sigma_1^2}{\nu^2}) + \log(1/\varepsilon)}{\log(1/\rho)}\right\rceil\,.
\end{align*}
For $\varepsilon \less \min\{\frac{\nu^2}{\sigma_1^2+\nu^2}, 1/2\}$, this reduces to 
\begin{align*}
    T = \mathcal{O}\left(\frac{\log(1/\varepsilon)}{\log(1/\rho)}\right)\,.
\end{align*}
Thus, we obtain the total time complexity
\begin{align*}
    C_\varepsilon = \mathcal{O}\left(\log (d_e/\rho) \cdot (\frac{d_e^2 \log^2\!d_e}{\rho^2} \, d + nd \log(d_e / \rho)) + nd\, \frac{\log(1/\varepsilon)}{\log(1/\rho)}\right)\,,
\end{align*}
which is the claimed result.

\section{Proofs of concentration inequalities}

\subsection{Gaussian concentration over ellipsoids -- Proof of Theorem~\ref{TheoremGaussianConcentrationEllipsoids}}
\label{ProofGaussianConcentrationEllipsoids}

Let $\rho > 0$ and $m \gre \frac{d_e}{\rho}$. Let $S \in \real^{m \times n}$ be a random matrix with i.i.d.~entries $\mathcal{N}(0,1/m)$. We aim to control the quantities
\begin{align*}
    & \gamma_1 = \sup_{\|x\|=1} 1 + \langle x, D (U^\top S^\top S U - I_d) D x \rangle \\
    & \gamma_d = \inf_{\|x\|=1} 1 + \langle x, D (U^\top S^\top S U - I_d) D x \rangle\,.
\end{align*}

\subsubsection*{Upper bound on the largest eigenvalue $\gamma_1$}

We introduce the re-scaled matrix $\bar{D} = \frac{D}{\|D\|_2}$, so that $\|\bar{D}\|_F^2 = d_e$ and $\|\bar{D}\|_2 = 1$. We have that
\begin{align*}
    \frac{\gamma_1 - 1}{\|D\|_2^2} \overset{\mathrm{d}}{=} \sup_{\|x\|=1} \langle x, \bar{D}(\frac{1}{m} G^\top G - I ) \bar{D} x\rangle &= \sup_{\|x\|=1} \frac{1}{m} \|G\bar{D}x\|^2 - \|\bar{D}x\|^2\\
    &= \frac{2}{m} \sup_{z \in \mathcal{C}} \sup_{u \in \real^m} u^\top G z + \psi(u,z)\,,
\end{align*}
where we introduced the random matrix $G \in \real^{m \times d}$ with i.i.d.~Gaussian entries $\mathcal{N}(0,1)$ and the first equality holds since $SU \overset{\mathrm{d}}{=} \frac{1}{\sqrt{m}} G$. We also used the notations $\mathcal{C} = \left\{\bar{D}x \mid \|x\|=1\right\}$ and $\psi(u,z) \defn -\frac{1}{2} (\|u\|^2 + m \|z\|^2)$. We introduce the auxiliary random variable 
\begin{align*}
    Y \defn \frac{2}{m} \, \sup_{z \in \mathcal{C}} \sup_{u \in \real^m} \|z\| g^\top u + \|u\| h^\top z + \psi(u,z)\,,
\end{align*}
where $g \in \real^m$ and $h \in \real^d$ are random vectors with i.i.d.~entries $\mathcal{N}(0,1)$. Using Theorem~\ref{TheoremMaxMax} (see Appendix~\ref{SectionTheoremMaxMax}), it holds that for any $c \in \real$, 
\begin{align}
\label{EqnTightUpperBound}
    \mathbb{P}\left(\frac{\gamma_1 - 1}{\|D\|^2_2} \gre c\right) \less 2 \,\mathbb{P}(Y \gre c)\,.
\end{align}
Consequently, it suffices to control the upper tail of $Y$ in order to control that of $\gamma_1$. First, we recall a few basic facts on the concentration of Gaussian random vectors (see, for instance, Theorems~3.1.1 and~6.3.2 in~\cite{vershynin2018high}). That is, for any $\eta > 0$, the following event holds with probability at least $1-4 e^{-m \rho \eta / 2}$,
\begin{align}
\label{EqnGaussianVectorConcentrationBounds1}
    \mathcal{E}_\eta \defn \left\{|\|g\|-\sqrt{m}| \less \sqrt{m \eta \rho} \,,\qquad |\|g\|^2 - m| \less m \sqrt{\eta \rho}\,, \qquad  \|\bar{D}h\|  \less \sqrt{m \rho} (1 +\sqrt{\eta})\right\} \,,
\end{align}
On the event $\mathcal{E}_\eta$, we have
\begin{align*}
    Y &= \frac{2}{m} \sup_{z \in \mathcal{C}} \sup_{u \in \real^m} \|z\| g^\top u + \|u\| h^\top z - \frac{1}{2} \|u\|^2 - \frac{m}{2} \|z\|^2\\
    &\overset{(i)}{=} \frac{2}{m} \sup_{z \in \mathcal{C}} \sup_{t \gre 0} \, t\, \|z\| \|g\| + t\, h^\top z - \frac{1}{2} t^2 - \frac{m}{2} \|z\|^2\\
    &\overset{(ii)}{\less} \frac{2}{m} \sup_{z \in \mathcal{C}} \sup_{t \in \real} \, t ( \|z\| \|g\| + |h^\top z|) - \frac{1}{2} t^2 - \frac{m}{2} \|z\|^2\\
    &\overset{(iii)}{\less} \frac{2}{m} \sup_{z \in \mathcal{C}} \frac{\|z\|^2}{2} |\|g\|^2-m| + \frac{1}{2} |h^\top z|^2 + \|z\| \|g\| |h^\top z| \\
    & \overset{(iv)}{\less} \frac{2}{m} \sup_{z \in \mathcal{C}} \frac{|\|g\|^2-m|}{2} + \frac{1}{2} |h^\top z|^2 + \|g\| |h^\top z| \\
    & \overset{(v)}{=} \frac{|\|g\|^2-m|}{m} + \frac{\|\bar{D}h\|^2}{m} + \frac{2 \|\bar{D}h\| \|g\|}{m}\\
    &\overset{(vi)}{\less}\sqrt{\rho \eta} + \rho (1+\sqrt{\eta})^2 + 2 \sqrt{\rho} (1+\sqrt{\eta}) (1+\sqrt{\rho \eta})\\
    & = \rho (1+\sqrt{\eta}) (1 + 3\sqrt{\eta})+ 2 \sqrt{\rho} (1+\frac{3}{2}\sqrt{\eta})\\
    & \less \left(1 + \sqrt{\rho c_\eta}\right)^2 - 1\,,
\end{align*}
where $c_\eta \defn (1+3\sqrt{\eta})^2$. In equality (i), we used the fact that for a vector $u$ with fixed norm $\|u\|=t$, the maximum of $g^\top u$ is equal to $\|g\| t$. In inequality (ii), we bounded $h^\top z$ by $|h^\top z|$ and then relaxed the constraint $t \gre 0$ to $t \in \real$. In inequality (iii), we plugged-in the value of the maximizer $t^* = \|z\|\|g\| + |h^\top z|$. In inequality (iv), we used the fact that for $z \in \mathcal{C}$, $\|z\| \less 1$. In (v), we used the fact that $\sup_{z \in \mathcal{C}} |h^\top z| = \|\bar{D} h\|$. In (vi), we used that, on the event $\mathcal{E}_\eta$, we have $\frac{|\|g\|^2-m|}{m} \less \sqrt{\eta m}$, $\|\bar{D}h\| \less \sqrt{m \rho}(1+\sqrt{\eta})$ and $\|g\| \less \sqrt{m}(1+\sqrt{\eta \rho})$. Consequently, we have that
\begin{align*}
    \mathbb{P}\left[\frac{\gamma_1-1}{\|D\|_2^2} \gre (1+\sqrt{\rho c_\eta})^2 - 1\right] &\less 2 \, \mathbb{P}\left[Y \gre (1+\sqrt{\rho c_\eta})^2 - 1\right]\\ &\less 2 (1-\mathbb{P}[\mathcal{E}_\eta])\\ 
    &\less 8 \cdot e^{-m \rho \eta /2}\,,
\end{align*}
which is the claimed upper bound~\eqref{eqngaussiancontrationbounds} on $\gamma_1$.

\subsubsection*{Controlling the smallest eigenvalue $\gamma_d$}

Here we assume that $\rho \in (0,0.18]$ and $\eta \in (0,0.01]$. We make this assumption in order to provide explicit and simple statements.

We consider the same definitions $\bar D$, $\mathcal{C}, \varphi$ and $\mathcal{E}_\eta$ introduced in the proof of the upper bound on $\gamma_1$. We have that
\begin{align*}
    \frac{\gamma_d - 1}{\|D\|_2^2} \overset{\mathrm{d}}{=} \inf_{\|x\|=1} \langle x, \bar{D}(\frac{1}{m} G^\top G - I ) \bar{D} x\rangle &= \inf_{\|x\|=1} \frac{1}{m} \|G\bar{D}x\|^2 - \|\bar{D}x\|^2\\
    &= \frac{2}{m} \inf_{z \in \mathcal{C}} \sup_{u \in \real^m} u^\top G z + \psi(u,z)\,.
\end{align*}
We introduce the auxiliary random variable 
\begin{align*}
    Y \defn \frac{2}{m} \, \inf_{z \in \mathcal{C}} \sup_{u \in \real^m} \|z\| g^\top u + \|u\| h^\top z + \psi(u,z)\,,
\end{align*}
where $g \in \real^m$ and $h \in \real^d$ are random vectors with i.i.d.~entries $\mathcal{N}(0,1)$. Using Theorem II.1 from~\cite{thrampoulidis2014tight}, it holds that for any $c \in \real$, 
\begin{align}
\label{EqnTightLowerBound}
    \mathbb{P}(\frac{\gamma_d - 1}{\|D\|_2^2} < c) \less 2 \,\mathbb{P}(Y < c)\,.
\end{align}
Consequently, it suffices to control the lower tail of $Y$ in order to control that of $\gamma_d$. It holds that
\begin{align*}
    Y &= \frac{2}{m} \inf_{z \in \mathcal{C}} \sup_{u \in \real^m} \|z\| g^\top u + \|u\| h^\top z - \frac{1}{2} \|u\|^2 - \frac{m}{2} \|z\|^2\\
    &= \frac{2}{m} \inf_{z \in \mathcal{C}} \sup_{t \gre 0} \, t\, \|z\| \|g\| + t\, h^\top z - \frac{1}{2} t^2 - \frac{m}{2} \|z\|^2\\
    &= \inf_{z \in \mathcal{C}} \begin{cases} -\|z\|^2\,,\quad \textrm{if}\,\,\|z\|\|g\| + h^\top z \less 0\\
    \frac{\|z\|^2}{m}(\|g\|^2 - m) + \frac{(h^\top z)^2}{m} + \frac{2}{m} \|z\|\|g\| (h^\top z)\,,\quad \textrm{otherwise.}\end{cases}
\end{align*}
Define 
\begin{align*}
    & Y_1 \defn \inf_{\substack{z \in \mathcal{C};\\\|z\| \|g\| + h^\top z \less 0}} -\|z\|^2\,,\\ 
    & Y_2 \defn \inf_{\substack{z \in \mathcal{C}\\\|z\|\|g\|+h^\top z \gre 0}} \frac{\|z\|^2}{m}(\|g\|^2 - m) + \frac{(h^\top z)^2}{m} + \frac{2}{m} \|z\|\|g\| (h^\top z)\,,
\end{align*}
so that $Y = \min\{Y_1, Y_2\}$. For any $z \in \mathcal{C}$, it holds that $h^\top z \gre - \|Dh\|$, and consequently
\begin{align*}
    Y_1 \gre \inf_{\substack{z \in \mathcal{C};\\\|z\| \|g\| \less \|\bar{D}h\|}} -\|z\|^2
    \gre -\frac{\|\bar{D}h\|^2}{\|g\|^2}\,.
\end{align*}
Hence, conditional on the event $\mathcal{E}_\eta$, we have
\begin{align*}
    Y_1 \gre -\rho \left(\frac{1+\sqrt{\eta}}{1-\sqrt{\rho\eta}}\right)^2\,,
\end{align*}
On the other hand, we have
\begin{align*}
    Y_2 &\gre - \frac{1}{m} |\|g\|^2 - m| + \inf_{z \in \mathcal{C}} \left\{ \frac{(h^\top z)^2}{m} - \frac{2}{m} \|g\| |h^\top z| \right\}\\
    &\gre - \frac{1}{m} |\|g\|^2 - m| + \inf_{\|x\|=1} \left\{ \frac{\langle \bar{D}h, x \rangle^2}{m} - \frac{2}{m} \|g\| |\langle \bar{D}h, x\rangle| \right\}\\
    &= - \frac{1}{m} |\|g\|^2 - m| + \frac{2}{m} \inf_{0 \less t \less \|\bar{D}h\|} \left\{ \frac{t^2}{2} - \|g\| t \right\}\,,
\end{align*}
where, in the first inequality, we relaxed the constraint set by removing the constraint $\|z\|\|g\| + h^\top z \gre 0$ and we used the fact that $\|z\| \less 1$. In the second inequality, we used the change of variable $z = \bar{D}x$ with $\|x\|=1$. In the third inequality, we used the fact that $|\langle \bar{D}h,x\rangle| \less \|\bar{D}h\|$ and used the change of variable $|\langle \bar{D}h, x\rangle| = t$ with $t \in [0, \|\bar{D}h\|]$. On the event $\mathcal{E}_\eta$, it follows that
\begin{align*}
    Y_2 &\gre \rho (1-\eta) - 2\sqrt{\rho} (1+\frac{3}{2}\sqrt{\eta})\\
    & \underset{(i)}{\gre} (1+3\sqrt{\eta})^2 \rho - 2 \sqrt{\rho} (1+3\sqrt{\eta})\\
    & = (1-\sqrt{c_\eta \rho})^2 - 1\,,
\end{align*}
One can verify that inequality (i) is equivalent to $\sqrt{\rho} \less \frac{1}{2 + \frac{10 \sqrt{\eta}}{3}}$, which always holds under the assumption that $\rho \less 0.18$ and $\eta \less 0.01$. Then, combining the respective lower bounds on $Y_1$ and $Y_2$, we obtain that
\begin{align*}
    Y &\gre \min\left\{ -\rho \left(\frac{1+\sqrt{\eta}}{1-\sqrt{\eta}}\right)^2,  (1-\sqrt{c_\eta \rho})^2 - 1\right\}\\
    &\gre (1-\sqrt{c_\eta \rho})^2 - 1\,,
\end{align*}
One can verify that the last inequality is equivalent to 
\begin{align*}
    \sqrt{\rho} \less \frac{2 (1+3\sqrt{\eta})}{(1+3\sqrt{\eta})^2 + \left(\frac{1+\sqrt{\eta}}{1-\sqrt{\eta}}\right)^2}\,,
\end{align*}
which always holds the assumption that $\rho \less 0.18$ and $\eta \less 0.01$.

Thus, we have proved the claimed lower bound on $\gamma_1$.

\subsubsection{A new Gaussian comparison inequality}
\label{SectionTheoremMaxMax}

We start with the following well-known comparison inequality, which was first derived in~\cite{gordon1985some}.

\btheos[Gordon's Gaussian comparison theorem]

Let $I, J \in \mathbb{N}^*$, and $\{X_{ij}\}$, $\{Y_{ij}\}$ be two centered Gaussian processed indexed on $I \times J$, such that for any $i, l \in I$ with $i \neq l$ and $j, k \in J$,
\begin{align*}
\begin{cases}
    \Exs X_{ij}^2 = \Exs Y_{ij}^2\\
    \Exs X_{ij} X_{ik} \gre \Exs Y_{ij} Y_{ik}\\
    \Exs X_{ij} X_{lk} \less \Exs Y_{ij} Y_{lk}\,.
\end{cases}
\end{align*}
Then, for any $\{\lambda_{ij}\} \in \real^{I \times J}$, we have
\begin{align*}
    \mathbb{P}\left(\bigcap_{i=1}^I \,\bigcup_{j=1}^J \,[Y_{ij} \gre \lambda_{ij}]\right) \gre \mathbb{P}\left(\bigcap_{i=1}^I\, \bigcup_{j=1}^J \,[X_{ij} \gre \lambda_{ij}]\right)
\end{align*}
\etheos
\noindent Our next result is a consequence of Gordon's comparison inequality, and appears to be new. More specifically, it can be seen as a variant of the Sudakov-Fernique's inequality (see, for instance, Theorem~7.2.11 in~\cite{vershynin2018high}).
\btheos
\label{TheoremMaxMax}
Let $S_1 \subset \real^n$ and $S_2 \subset \real^m$ be non-empty sets, and $\psi : S_1 \times S_2 \to \real$ be a continuous function. Then, for any $c \in \real$,
\begin{align*}
    \mathbb{P}\!\left(\sup_{(x,y) \in S_1 \times S_2} y^\top Gx + \psi(x,y) \gre c\right) \, \less \, 2 \, \mathbb{P}\!\left(\sup_{(x,y) \in S_1 \times S_2} \|x\| g^\top y + \|y\| h^\top x + \psi(x,y) \gre c\right)\,,
\end{align*}
\etheos
\spro
The proof relies on several intermediate results, and is deferred to Section~\ref{ProofTheoremMaxMax}.
\fpro 
\blems
\label{LemmaConvexSudakov}
Let $G \in \real^{m \times n}$, $Z \in \real$, $g \in \real^m$ and $h \in \real^n$ have independent standard Gaussian entries. Let $I_1 \subset \real^n$ and $I_2 \subset \real^m$ be finite sets, and $\psi$ be a function defined over $I_1 \times I_2$. Then, for any $c \in \real$, we have
\begin{align*}
    \mathbb{P}\left( \max_{(x,y) \in I_1 \times I_2} y^\top G x + Z \|x\| \|y\| + \psi(x,y) \gre c \right) \less \mathbb{P}\left( \max_{(x,y) \in I_1 \times I_2} \|x\| g^\top y + \|y\| h^\top x + \psi(x,y) \gre c \right)\,.
\end{align*}
\elems
\spro 
We introduce two Gaussian processes $X$ and $Y$ indexed over $I_1 \times I_2$, defined as
\begin{align*}
    X_{xy} = \|x\| g^\top y + \|y\| h^\top x\,,\qquad Y_{xy} = y^\top G x + Z \|x\| \|y\|\,,
\end{align*}
for all $(x,y) \in I_1 \times I_2$. It holds that $\Exs X_{xy} = \Exs Y_{xy} = 0$, $\Exs X^2_{xy} = 2 \|x\|^2\|y\|^2 = \Exs Y^2_{xy}$, and 
\begin{align*}
    & \Exs \!\left[X_{xy} X_{x^\prime y^\prime}\right] = \|x\| \|x^\prime\| \, y^\top y^\prime + \|y\| \|y^\prime\| \, x^\top x^\prime\,, \\
    & \Exs \!\left[Y_{xy} Y_{x^\prime y^\prime}\right] = \|x\| \, \|x^\prime\| \, \|y\| \, \|y^\prime\| + x^\top x^\prime \, y^\top y^\prime\,.
\end{align*}
Consequently, we have
\begin{align*}
    \Exs \!\left[Y_{xy} Y_{x^\prime y^\prime}\right] - \Exs \!\left[X_{xy} X_{x^\prime y^\prime}\right] &= \left(\|x\|\,\|x^\prime\| - x^\top x^\prime \right) \left(\|y\|\,\|y^\prime\| - y^\top y^\prime \right)\\
    & \gre 0\,.
\end{align*}
Therefore, applying Gordon's comparison theorem with $I=I_1 \times I_2$, $J$ being any finite set, and $\lambda_{xy} = \psi(x,y)-c$, we obtain that 
\begin{align*}
    \mathbb{P}\!\left(\min_{(x,y) \in I_1 \times I_2} y^\top Gx + Z \|x\| \|y\| - \psi(x,y) \gre -c\right) \,\gre\, \mathbb{P}\!\left(\min_{(x,y) \in I_1 \times I_2} \|x\| g^\top y + \|y\| h^\top x - \psi(x,y) \gre -c\right)\,.
\end{align*}
Using the symmetry of the Gaussian distribution, it follows that
\begin{align*}
    \mathbb{P}\!\left(\max_{(x,y) \in I_1 \times I_2} y^\top Gx + Z \|x\| \|y\| + \psi(x,y) \less c\right) \,\gre\, \mathbb{P}\!\left(\max_{(x,y) \in I_1 \times I_2} \|x\| g^\top y + \|y\| h^\top x + \psi(x,y) \less c\right)\,,
\end{align*}
and consequently,
\begin{align*}
    \mathbb{P}\!\left(\max_{(x,y) \in I_1 \times I_2} y^\top Gx + Z \|x\| \|y\| + \psi(x,y) \gre c\right) \, \less \, \mathbb{P}\!\left(\max_{(x,y) \in I_1 \times I_2} \|x\| g^\top y + \|y\| h^\top x + \psi(x,y) \gre c\right)\,,
\end{align*}
\fpro 
\bcors
\label{CorollaryConvexSudakov}
Let $S_1 \subset \real^n$ and $S_2 \subset \real^m$ be non-empty sets, and $\psi : S_1 \times S_2 \to \real$ be a continuous function. Then, for any $c \in \real$,
\begin{align*}
    \mathbb{P}\!\left(\sup_{(x,y) \in S_1 \times S_2} y^\top Gx + Z \|x\| \|y\| + \psi(x,y) \gre c\right) \less \mathbb{P}\!\left(\sup_{(x,y) \in S_1 \times S_2} \|x\| g^\top y + \|y\| h^\top x + \psi(x,y) \gre c\right)\,,
\end{align*}
\ecors
\spro 
According to Lemma~\ref{LemmaConvexSudakov}, the result is true if $S_1$ and $S_2$ are finite. By monotone convergence, it is immediate to extend it to countable sets. By density arguments and monotone convergence, it also follows for any sets $S_1$ and $S_2$.
\fpro

\subsubsection{Proof of Theorem~\ref{TheoremMaxMax}}
\label{ProofTheoremMaxMax}

We define $f_1(x,y) = y^\top Gx + \psi(x,y)$ and $f_2(x,y) = y^\top Gx + Z \|x\| \|y\| + \psi(x,y)$. If $Z > 0$, then $f_1 \less f_2$ and $\sup_{x,y} f_1(x,y) \less \sup_{x,y} f_2(x,y)$. Thus,
\begin{align*}
    \mathbb{P}\!\left( \sup_{(x,y) \in S_1 \times S_2} f_1(x,y) \gre c, \quad Z > 0\right) \, \less \, \mathbb{P}\!\left( \sup_{(x,y) \in S_1 \times S_2} f_2(x,y) \gre c\right)\,.
\end{align*}
From Corollary~\ref{CorollaryConvexSudakov}, we know that 
\begin{align*}
    \mathbb{P}\!\left( \sup_{(x,y) \in S_1 \times S_2} f_2(x,y) \gre c\right) \less \mathbb{P}\!\left(\sup_{(x,y) \in S_1 \times S_2} \|x\| g^\top y + \|y\| h^\top x + \psi(x,y) \gre c\right)\,.
\end{align*}
Consequently, using the independence of $f_1$ and $Z$, we get
\begin{align*}
    \frac{1}{2} \,\mathbb{P}\!\left( \sup_{(x,y) \in S_1 \times S_2} f_1(x,y) \gre c \right) &= \, \mathbb{P}\!\left( \sup_{(x,y) \in S_1 \times S_2} f_1(x,y) \gre c, \quad Z > 0\right)\\
    & \less \, \mathbb{P}\!\left(\sup_{(x,y) \in S_1 \times S_2} \|x\| g^\top y + \|y\| h^\top x + \psi(x,y) \gre c\right)\,,
\end{align*}
which yields the claim.
\qed

\subsection{SRHT matrices -- matrix deviation inequalities over ellipsoids}
\label{SectionProofSRHTMatrices}

\subsection{Preliminaries}

Let $S \in \real^{m \times n}$ be a SRHT matrix, that is, $S = R H \text{diag}(\varepsilon)$ where $R$ is a row-subsampling matrix of size $m \times n$, $H$ is the normalized Walsh-Hadamard transform of size $n \times n$ and $\varepsilon$ is a vector of $n$ independent Rademacher variables. We introduce the scaled diagonal matrix $\bar{D} = \frac{D}{\|D\|_2}$. Note that $\|\bar{D}\|_F^2 = d_e$ and $\|\bar{D}\|_2 = 1$.
\blems 
\label{LemmaRowNormsScaled}
Let $e_j$ be the $j$-th vector of the canonical basis in $\real^n$. Then,
\begin{align}
    \mathbb{P}\left\{ \max_{j=1,\hdots,n} \|e_j^\top H \text{diag}(\varepsilon) U \bar{D}\| \gre \sqrt{\frac{d_e}{n}} + \sqrt{\frac{8 \log(\beta n)}{n}} \right\} \less \frac{1}{\beta}\,.
\end{align}
\elems  
\spro
We fix a row index $j \in \{1,\hdots,n\}$, and define the function
\begin{align*}
    f(x) \defn \|e_j^\top H \text{diag}(x) U \bar{D}\| = \|x^\top E U \bar{D}\|\,,    
\end{align*}
where $E \defn \text{diag}(e_j^\top H)$. Each entry of $E$ has magnitude $n^{-\frac{1}{2}}$. The function $f$ is convex, and its Lipschitz constant is upper bounded as follows,
\begin{align*}
    |f(x)-f(y)| \less \|(x-y)^\top E V \bar{D}\| \less \|x-y\|\, \|E\|_2\, \|V\|_2\, \|\bar{D}\|_2 = \frac{1}{\sqrt{n}} \|x-y\|\,.
\end{align*}
For a Rademacher vector $\varepsilon$, we have
\begin{align*}
    \Exs \, f(\varepsilon) \less \sqrt{\Exs \, f(\varepsilon)^2} = \|EU\bar{D}\|_F \less \|EU\|_2 \, \|\bar{D}\|_F =  \sqrt{\frac{d_e}{n}}\,.
\end{align*}
Applying Lipschitz concentration results for Rademacher variables, we obtain
\begin{align*}
    \mathbb{P}\left\{\|e_j^\top H \text{diag}(\varepsilon) U \bar{D}\| \gre \sqrt{\frac{d_e}{n}} + \sqrt{\frac{8 \log(\beta n)}{n}} \right\} \less \frac{1}{n \beta}\,.
\end{align*}
Finally, taking a union bound over $j \in \{1,\hdots,n\}$, we obtain the claimed result.
\fpro
\btheos[Matrix Bernstein] 
\label{ThmBernstein}
Let $\mathcal{X} = \{X_1, \hdots, X_n\}$ be a finite set of squared matrices with dimension $d$. Fix a dimension $m$, and suppose that there exists a positive semi-definite matrix $V$ and a real number $K > 0$ such that $\Exs[X_I] = 0$, $\Exs[X_I^2] \preceq V$, and $\|X_I\|_2 \less K$ almost surely, where $I$ is a uniformly random index over $\{1,\hdots,n\}$. Let $T$ be a subset of $\{1,\hdots,n\}$ with $m$ indices drawn uniformly at random without replacement. Then, for any $t \gre \sqrt{m \|V\|_2} + K/3$, we have
\begin{align*}
    \mathbb{P} \left\{ \Big\| \sum_{i \in T} X_i \Big\|_2 \gre t \right\} \, \less 8\cdot d_e \cdot \exp \left( -\frac{t^2/2}{m \|V\|_2 + K t /3} \right)\,,
\end{align*}
where $d_e \defn \text{tr}(V) / \|V\|$ is the intrinsic dimension of the matrix $V$.
\etheos 
\spro 
We denote $S_T \defn \sum_{i \in T} X_i$. Fix $\theta > 0$, define $\psi(t)=e^{\theta t} - \theta t - 1$, and use the Laplace matrix transform method (e.g., Proposition~7.4.1 in~\cite{tropp2015introduction}) to obtain
\begin{align*}
    \mathbb{P} \left\{ \lambda_\text{max}(S_T) \gre t \right\} \, &\less \, \frac{1}{\psi(t)} \, \Exs \, \text{tr} \,\psi(S_T)\\
    & = \frac{1}{e^{\theta t} - \theta t - 1} \Exs \, \text{tr}\left(e^{\theta S_T} - I\right)\,,
\end{align*}
and the last equality holds due to the fact that $\Exs \, S_T = m \,\Exs \, X_I = 0$. Let $T^\prime = \{i_1, \hdots, i_m\}$ be a subset of $\{1, \hdots, n\}$, drawn uniformly at random \textit{with} replacement. In particular, the indices of $T^\prime$ are independent random variables, and so are the matrices $\{X_{i_j}\}_{j=1}^m$. Write $S_{T^\prime} \defn \sum_{j=1}^m X_{i_j}$. Gross and Nesme~\cite{gross2010note} have shown that for any $\theta > 0$,
\begin{align*}
    \Exs \, \text{tr} \exp\left( \theta S_T \right) \, \less \, \Exs \, \text{tr} \exp \left( \theta S_{T^\prime} \right)\,.
\end{align*}
As a consequence of Lieb's inequality (e.g., Lemma~3.4 in~\cite{tropp2015introduction}), it holds that
\begin{align*}
    \Exs \, \text{tr} \exp \left( \theta S_{T^\prime} \right) \, \less \text{tr} \exp \left( \sum_{j=1}^m \log \, \Exs \, e^{\theta X_{i_j}} \right) = \, \text{tr} \exp \left( m \, \log \, \Exs \, e^{\theta X_I} \right)\,.
\end{align*}
Thus, it remains to bound $\Exs \, e^{\theta X_I}$. By assumption, $\Exs[X_I] = 0$ and $\|X_I\|_2 \less K$ almost surely. Then, using Lemma~5.4.10 from~\cite{vershynin2018high}, we get $\Exs \, e^{\theta X_I} \preceq \exp\!\left(g(\theta) \, \Exs \, X_I^2 \right)$, for any $|\theta| < 3/K$ and where $g(\theta) = \frac{\theta^2/2}{1-|\theta| K /3}$. By monotonicity of the logarithm, $m \cdot \log \Exs \, e^{\theta X_I} \preceq m \cdot g(\theta) \, \Exs \,X_I^2$. By assumption, $\Exs \, X_I^2 \preceq V$ and thus, $m \cdot \log \Exs \, e^{\theta X_I} \preceq m \cdot g(\theta) \, V$. By monotonicity of the trace exponential, it follows that $\text{tr} \exp\!\left( m \, \log \, \Exs \, e^{\theta X_I} \right) \less \text{tr} \exp \left(m \,g(\theta)\, V\right)$, and further,
\begin{align*}
    \mathbb{P} \left\{ \lambda_\text{max}(S_T) \gre t \right\} \, \less \, \frac{1}{e^{\theta t} - \theta t - 1} \text{tr} \left( e^{m \,g(\theta)\, V} - I \right) = \frac{1}{e^{\theta t} - \theta t - 1} \,\text{tr} \, \varphi(m\,g(\theta)\,V) \,,
\end{align*}
where $\varphi(a) = e^a - 1$. The function $\varphi$ is convex, and the matrix $m \,g(\theta)\, V$ is positive semidefinite. Therefore, we can apply Lemma~7.5.1 from~\cite{tropp2015introduction} and obtain
\begin{align*}
    \text{tr} \, \varphi(m\,g(\theta)\,V) \less d_e \cdot \varphi(m\,g(\theta)\,\|V\|_2) \less d_e \cdot e^{m\,g(\theta)\,\|V\|_2}\,,
\end{align*}
which further implies that
\begin{align*}
    \mathbb{P} \left\{ \lambda_\text{max}(S_T) \gre t \right\} \, \less \, d_e \cdot \frac{e^{\theta t}}{e^{\theta t} - \theta t - 1} \cdot e^{-\theta t + m \,g(\theta) \cdot \|V\|_2} \less d_e \cdot \left(1+\frac{3}{\theta^2 t^2}\right) \cdot e^{-\theta t + m\,g(\theta) \cdot \|V\|_2}\,.
\end{align*}
For the last inequality, we used the fact that $\frac{e^a}{e^a - a - 1} = 1 + \frac{1+a}{e^a - a -1} \less 1 + \frac{3}{a^2}$ for all $a \gre 0$. Picking $\theta = t / (m\,\|V\|_2 + Kt/3)$, we obtain 
\begin{align*}
    \mathbb{P} \left\{ \lambda_\text{max}(S_T) \gre t \right\} \, \less d_e \cdot \left(1+ 3\cdot \frac{(m \|V\|_2 + Kt/3)^2}{t^4}\right) \cdot \exp \left( -\frac{t^2/2}{m \|V\|_2 + K t /3} \right)\,.
\end{align*}
Under the assumption $t \gre \sqrt{m \|V\|_2} + K / 3$, the parenthesis in the above right-hand side is bounded by four, which results in 
\begin{align*}
    \mathbb{P} \left\{ \lambda_\text{max}(S_T) \gre t \right\} \, \less 4\cdot d_e \cdot \exp \left( -\frac{t^2/2}{m \|V\|_2 + K t /3} \right)\,.
\end{align*}
Repeating the argument for $-S_T$ and combining the two bounds, we obtain the claimed result.
\fpro

\subsubsection{Proof of Theorem~\ref{TheoremMainBoundSRHT}}
\label{ProofTheoremMainBoundSRHT}

We write $v_j \defn \sqrt{\frac{n}{m}}\,w_j$, where $w_j = e_j^\top H \text{diag}(\varepsilon) U \bar{D}$, and $\varepsilon \in \{\pm 1\}^n$ is a fixed vector. We denote 
\begin{align*}
    \gamma \defn \max\left\{\max_{j=1,\hdots,n} \|v_j\|, \, m^{-\frac{1}{2}} \right\} \qquad \text{and} \qquad X_i \defn v_i v_i^\top - \frac{1}{m} \bar{D}^2\,.
\end{align*} 
Let $I$ be a uniformly random index over $\{1,\hdots,n\}$. We have 
\begin{align*}
    \Exs[X_I] = \frac{n}{m} \Exs[w_I w_I^\top] - \frac{1}{m} \bar{D}^2 &= \frac{n}{m} \left(\frac{1}{n}\sum_{i=1}^n \bar{D}U^\top \text{diag}(\varepsilon) H e_i e_i^\top H \text{diag}(\varepsilon) U \bar{D}\right) - \frac{1}{m} \bar{D}^2\\
    &= \frac{1}{m} \bar{D}U^\top \text{diag}(\varepsilon) H \underbrace{\sum_{i=1}^n e_i e_i^\top}_{= I} H \text{diag}(\varepsilon) U \bar{D} - \frac{1}{m} \bar{D}^2\\
    &= 0\,.
\end{align*}
The last equality holds due to the fact that $H^2 = I$, $\text{diag}(\varepsilon)^2 = I$ and $U^\top U = I$. Further, $\|v_I\|^2 \less \gamma^2$ a.s., so that $\|v_I\|^2 v_I v_I^\top \preceq \gamma^2 v_I v_I^\top$ a.s., and consequently, $\Exs\left[ \|v_I\|^2 v_I v_I^\top \right] \preceq \gamma^2 \cdot \Exs[v_I v_I^\top]$. Thus,
\begin{align*}
    \Exs\left[X_I^2\right] &= \Exs\left[ \|v_I\|^2 v_I v_I^\top \right] - \frac{2}{m} \bar{D}^4 + \frac{1}{m^2} \bar{D}^4\\
    & \less \frac{\gamma^2}{m} \bar{D}^2 - \frac{2}{m^2} \bar{D}^4 + \frac{1}{m^2} \bar{D}^4\\
    &= \gamma^2 \cdot \frac{1}{m} \bar{D}^2 - \frac{1}{m^2} \bar{D}^4\\
    & \preceq \frac{\gamma^2}{m} \bar{D}^2\,. 
\end{align*}
The first inequality holds due to the fact that $\Exs\left[v_I v_I^\top\right] = m^{-1} \bar{D}^2$. Further, we have 
\begin{align*}
    \|X_I\| = \|v_I v_I^\top - \frac{1}{m} \bar{D}^2\| \less \max\left\{ \max_{j=1,\hdots,n} \|v_j\|^2, m^{-1}  \right\} = \gamma^2\,. 
\end{align*}
Let $T$ be a subset of $m$ indices in $\{1, \hdots, n\}$ drawn uniformly at random, without replacement. Applying Theorem~\ref{ThmBernstein} with $V = m^{-1} \gamma^2 \bar{D}^2$ and using the scale invariance of the effective dimension, we obtain that for any $t \gre \gamma + \gamma^2/3$,
\begin{align*}
    \mathbb{P} \left\{ \Big\| \sum_{i \in T} X_i \Big\|_2 \gre t \right\} \,\less 8 d_e \cdot \exp \left( -\frac{t^2/2}{\gamma^2 (1 + t /3)} \right)\,.
\end{align*}
Suppose now that $\varepsilon$ is a vector of independent Rademacher variables. Note that $\sum_{i \in T} X_i \overset{\mathrm{d}}{=} \bar{D} U^\top (S^\top S - I) U \bar{D}$. From Lemma~\ref{LemmaRowNormsScaled}, we know that $\gamma \less \sigma \defn \sqrt{\frac{d_e}{m}} + \sqrt{\frac{8 \log(d_e n)}{m}}$ with probability at least $1-d_e^{-1}$. Consequently, with probability at least $1 - d_e^{-1} - 8d_e \cdot \exp \left( -\frac{t^2/2}{\sigma^2 (1 + t /3)} \right)$, for $t \gre \sigma \,(1+\sigma/3)$ we have
\begin{align}
\label{EqnIntermediateSRHT1}
    \Big\| \bar{D} U^\top (S^\top S - I) U \bar{D} \Big\|_2 \less t\,.
\end{align}
We set $t=\sigma \sqrt{8/3 \log d_e}$, and $\rho = \frac{d_e \log(d_e) C(n,d_e)}{m}$ where $C(n,d_e) = \frac{16}{3} \left(1 + \sqrt{\frac{8 \log(d_e n)}{d_e}}\right)^2$. We choose $m$ large enough so that $\rho \less \left(1 - (8/3 \log d_e)^{-\frac{1}{2}}\right)^2$. Then, we get that
\begin{align*}
    \mathbb{P} \left\{ \Big\| D U^\top (S^\top S - I) U D \Big\|_2 \gre \|D\|_2^2 \cdot \sqrt{\rho} \right\} \,\less \, \frac{9}{d_e}\,,
\end{align*}
which is the claimed result. 
\qed

\section{Proofs of auxiliary results}

\subsection{Proof of Lemma~\ref{lemmanewtondecrement}}
\label{prooflemmanewtondecrement}

Let $\{x_t\}$ be a sequence of iterates. Let $\overline U \,\overline \Sigma\, \overline V^\top$ be a singular value decomposition of $\overline A$. Denote $\overline S = \begin{bmatrix} S & 0 \\ 0 & I_d \end{bmatrix}$, so that $H_S = (SA)^\top SA + \nu^2 I_d = (\overline S \, \overline A)^\top (\overline S \, \overline A)$.

We have that $g_t = \overline A^\top \, \overline A (x_t - x^*)$ and thus,
\begin{align*}
    g_t^\top H_S^{-1} g_t &= \langle \overline A^\top\, \overline A (x_t-x^*), (\overline{A}^\top \overline S^\top \overline S \, \overline A)^{-1} \overline A^\top\, \overline A (x_t - x^*) \rangle\\
    &= \langle \overline A (x_t - x^*), \overline A (\overline{A}^\top \,\overline S^\top\, \overline S\, \overline A)^{-1} \overline A^\top\, \overline A (x_t - x^*) \rangle\\
    &= \langle \overline A (x_t - x^*), \overline U \,\overline \Sigma \,\overline V^\top (\overline V \,\overline \Sigma\, \overline U^\top\, \overline S^\top \,\overline S \,\overline U \, \overline \Sigma \,\overline V^\top )^{-1} \overline V \, \overline \Sigma \, \overline U^\top \, \overline A (x_t - x^*)\rangle\\
    &= \langle \overline A (x_t - x^*), \overline U (\overline U^\top\, \overline S^\top \,\overline S \,\overline U)^{-1} \overline U^\top \, \overline A (x_t - x^*)\rangle\\
    &= \langle \overline U^\top \, \overline A (x_t - x^*), (\overline U^\top\, \overline S^\top \,\overline S \,\overline U)^{-1} \overline U^\top \, \overline A (x_t - x^*)\rangle\,.
\end{align*}
Observing that $\overline U^\top\, \overline S^\top \,\overline S \,\overline U = C_S$, it follows that $\frac{1}{2} \, g_t^\top H_S^{-1} g_t = \frac{1}{2} \|C_S^{-\frac{1}{2}} \overline U^\top \,\overline A (x_t-x^*)\|^2 = r_t$, which concludes the proof.
\qed

\subsection{Proof of Lemma~\ref{lemmagaussiandetails}}
\label{prooflemmagaussiandetails}

Fix $\rho \less 0.18$ and $\eta \less 0.01$. Let $a \gre 1$ be some numerical constant, and assume that the event $\mathcal{E}_{\rho/a,\eta}$ holds. Then, we have that
\begin{align*}
    \sqrt{c_\text{gd}(\rho/a,\eta)} = \frac{2}{\sqrt{a}}\frac{\sqrt{\rho c_\eta}}{1+\frac{\rho c_\eta}{a}}\,,\qquad \sqrt{\frac{\sigma_\text{max}(C_S)}{\sigma_\text{min}(C_S)}} \less \frac{\sqrt{a} + \sqrt{\rho c_\eta}}{\sqrt{a}-\sqrt{\rho c_\eta}}\,.
\end{align*}
Using that $\sqrt{\rho c_\eta} \less \sqrt{0.18 \cdot 1.3^2} \less 0.56$ and $\rho c_\eta \less 0.31$, we obtain that
\begin{align*}
    \sqrt{c_\text{gd}(\rho/a,\eta)}  \cdot \sqrt{\frac{\sigma_\text{max}(C_S)}{\sigma_\text{min}(C_S)}} &\less \frac{1}{\sqrt{a}}\,\frac{1+\rho c_\eta}{1+\frac{\rho c_\eta}{a}}\, \frac{\sqrt{a} + \sqrt{\rho c_\eta}}{\sqrt{a}-\sqrt{\rho c_\eta}}\cdot \sqrt{c_\text{gd}(\rho,\eta)}\\
    &\less \frac{1.31}{\sqrt{a}} \cdot \frac{\sqrt{a} + 0.56}{\sqrt{a}-0.56} \cdot \sqrt{c_\text{gd}(\rho,\eta)}\,.
\end{align*}
The function $g: x \mapsto \frac{1.31}{\sqrt{x}} \cdot \frac{\sqrt{x} + 0.56}{\sqrt{x}-0.56}$ is decreasing on $(0.56^2, +\infty)$ and $g(5) \less 1$. Thus, for any $a \gre 5$, it holds that 
\begin{align*}
    c_\text{gd}(\rho/a,\eta)  \cdot \frac{\sigma_\text{max}(C_S)}{\sigma_\text{min}(C_S)} \less c_\text{gd}(\rho,\eta)\,,
\end{align*}
and this concludes the proof.
\qed

\subsection{Proof of Lemma~\ref{lemmasrhtdetails}}
\label{prooflemmasrhtdetails}

By definition, we have on the event $\mathcal{E}_{\rho/a_\rho}$ that
\begin{align*}
    \lambda_{\rho/a_\rho} \less \sigma_\text{min}(C_S) \less \sigma_\text{max}(C_S) \less \Lambda_{\rho/a_\rho}\,,
\end{align*}
where $\lambda_{\rho/a_\rho} = 1-\sqrt{\frac{\rho}{a_\rho}}$ and $\Lambda_{\rho/a_\rho} = 1+\sqrt{\frac{\rho}{a_\rho}}$, and $a_\rho = \frac{1+\sqrt{\rho}}{1-\sqrt{\rho}}$. It follows that
\begin{align*}
    \frac{\sigma_\text{max}(C_S)}{\sigma_\text{min}(C_S)} \less \frac{1+\sqrt{\frac{\rho}{a_\rho}}}{1-\sqrt{\frac{\rho}{a_\rho}}} = \frac{\sqrt{a_\rho} + \sqrt{\rho}}{\sqrt{a_\rho}-\sqrt{\rho}}\,.
\end{align*}
The function $x \mapsto \frac{x + \sqrt{\rho}}{x - \sqrt{\rho}}$ is decreasing on $[1,+\infty)$. Since $a_\rho > 1$, it follows that $f(a_\rho) < f(1)$, i.e., $\frac{\sqrt{a_\rho} + \sqrt{\rho}}{\sqrt{a_\rho}-\sqrt{\rho}} < \frac{1 + \sqrt{\rho}}{1-\sqrt{\rho}}$, i.e., $\frac{\sqrt{a_\rho} + \sqrt{\rho}}{\sqrt{a_\rho}-\sqrt{\rho}} < a_\rho$, which yields that
\begin{align*}
    \frac{\sigma_\text{max}(C_S)}{\sigma_\text{min}(C_S)} \less a_\rho\,.
\end{align*}
Regarding the second statement of Lemma~\ref{lemmasrhtdetails}, a simple calculation yields that $c_\text{gd}(\rho^\prime) = \rho^\prime$ for any $\rho^\prime \in (0,1)$. This further implies that $c_\text{gd}(\rho/a_\rho) = \frac{\rho}{a_\rho} = \frac{c_\text{gd}(\rho)}{a_\rho}$, which concludes the proof. 
\qed

\end{document}